RESEARCH ARTICLE

# E2E Process Automation Leveraging Generative AI and IDP-Based Automation Agent: A Case Study on Corporate Expense Processing

**Dr. Cheonsu Jeong[1,*], Seongmin Sim[1], Hyoyoung Cho[1], Sungsu Kim[1] and Byounggwan Shin[1]**

[1] *Hyper Automation Team, SAMSUNG SDS, Seoul, 05510, South Korea*

***Corresponding author:** Cheonsu Jeong. SAMSUNG SDS Email: csu.jeong@samsung.com

**Abstract:** This paper presents an intelligent work automation approach in the context of contemporary digital transformation by integrating generative AI and Intelligent Document Processing (IDP) technologies with an Automation Agent to realize End-to-End (E2E) automation of corporate financial expense processing tasks. While traditional Robotic Process Automation (RPA) has proven effective for repetitive, rule-based simple task automation, it faces limita-tions in handling unstructured data, exception management, and complex decision-making. This study designs and implements a four-stage integrated process comprising automatic recognition of supporting documents such as receipts via OCR/IDP, item classification based on a policy-driven database, intelligent exception handling sup-ported by generative AI (large language models, LLMs), and human-in-the-loop final decision-making with con-tinuous system learning through an Automation Agent. Applied to a major Korean enterprise (Company S), the system demonstrated quantitative benefits including over 80% reduction in processing time for paper receipt ex-pense tasks, decreased error rates, and improved compliance, as well as qualitative benefits such as enhanced ac-curacy and consistency, increased employee satisfaction, and data-driven decision support. Furthermore, the system embodies a virtuous cycle by learning from human judgments to progressively improve automatic exception han-dling capabilities. Empirically, this research confirms that the organic integration of generative AI, IDP, and Au-tomation Agents effectively overcomes the limitations of conventional automation and enables E2E automation of complex corporate processes. The study also discusses potential extensions to other domains such as accounting, human resources, and procurement, and proposes future directions for AI-driven hyper-automation development.

**Keywords:** Generative AI, IDP (Intelligent Document Processing), Automation Agent, E2E Automation, Corporate Expense Processing

## 1. Introduction

The acceleration of digital transformation has led companies to expand their interest and investment in automating business processes to secure a competitive edge and achieve sustainable growth. In particular, the automation of repetitive and time-consuming office tasks is recognized as a key factor directly influencing productivity improvement and cost reduction. Initially, Robotic Process Automation (RPA) was primarily used to automate rule-based repetitive tasks [1], but recent advancements in Artificial Intelligence (AI) technology, particularly generative AI and Intelligent Document Processing (IDP), have overcome the limitations of RPA and opened possibilities for automating more complex and intelligent tasks. Generative AI, which can create new content such as text, images, audio, and video based on extensive training data, has enabled users to easily utilize generative AI services [2, 3]. Specifically, generative AI chatbots have reached the level of analyzing human emotions and intentions to provide responses [4], and with the advent of Large Language Models (LLMs), there have been significant improvements in automated conversation generation and translation [5]. However, generative AI can also generate responses conflicting with the latest information and may have a lower understanding of new problems or domains due to relying on previously learned content [6]. To address this, solutions such as domain-specific fine-tuning of LLMs and using internal information to enhance reliability through methods like RAG are being explored [7]. IDP goes beyond Optical Character Recognition (OCR) by understanding the structure and context of documents to accurately extract and classify necessary information [8]. Additionally, IDP is faster, cheaper, and more accurate than humans reading documents and inputting data [9]. These technological advancements are providing new opportunities for automating business processes in companies. Furthermore, the growth of the IDP market is driven by the increasing adoption of intelligent automation technologies to enhance productivity and efficiency across organizations. Core functions such as marketing, human resources, finance, and analytics are leveraging IDP software to automate tasks like insight generation, document scanning, and candidate selection, thereby reducing workloads. This allows organizations to minimize repetitive and manual document processing tasks and focus on more strategic and value-added activities [10]. In particular, in corporate accounting tasks, generative AI contributes to improving efficiency and accuracy through automation and data analysis [11]. Among accounting tasks, financial expense processing is a representative repetitive business process that all organizations face. However, many companies still rely on inefficient methods such as manual processing and paper receipts for expense management. Manual expense management is prone to human errors, and complex processes like organizing paper receipts, filling out forms

The main reason for the delay in financial accounting tasks is the manual verification and review process for expense processing, which necessitates the introduction of automation to address this issue [9]. Many companies are actively seeking solutions to improve the efficiency of their expense processing processes to resolve such inefficiencies.

This study aims to address these issues by proposing a solution that combines generative AI and OCR/IDP technologies with Automation Agents to achieve End-to-End (E2E) automation for corporate expense management processes. Through real-world implementation cases, it seeks to validate the effectiveness and potential of this approach. This goes beyond the automation of simple repetitive tasks, as the system intelligently handles and learns from exception situations that require human judgment, marking a significant step toward true Hyperautomation.

The main objectives of this study are as follows: First, to understand the concepts and characteristics of generative AI, OCR/IDP, and Automation Agent technologies, and theoretically explore how these technologies can contribute to the automation of complex business processes, such as financial expense management, when combined. Second, to conduct an in-depth analysis of actual cases where end-to-end automation was implemented in financial expense management tasks using Automation Agent solutions. Third, to compare the quantitative and qualitative effects before and after the introduction of the proposed automation system, and through this, to evaluate the impact of generative AI and OCR/IDP-based Automation Agents on improving corporate productivity and enhancing operational efficiency.

This study is structured as follows. Chapter 1, the Introduction, presents the background and necessity of the research, as well as its objectives and scope. Chapter 2 examines the concepts and technological trends of generative AI, OCR/IDP, RPA, and Automation Agents, as well as E2E automation, and analyzes the characteristics of corporate expense processing tasks and the limitations of existing automation methods. Chapter 3 focuses on the automation of corporate expense processing using Automation Agents, detailing the problems of existing processes, the background and goals of introducing the automation system, and the four-stage automation implementation process from document recognition to user judgment reflection. Chapter 4 analyzes the quantitative and qualitative effects observed in actual cases, discusses the limitations encountered during implementation, and proposes future improvement plans. Chapter 5 summarizes the research findings, presents the implications and contributions of the study, and suggests directions for future related research.

# 2. Literature Review

## 2.1. Generative AI

Figures Generative AI (Gen AI) is a field of artificial intelligence that can generate new and original content by learning existing data [2, 12]. It can create various forms of data, such as text, images, audio, and video, and has shown remarkable achievements in recent years, particularly with advancements in deep learning technologies, especially the Transformer architecture and LLMs [13]. Generative AI distinguishes itself from Analytic AI, which merely recognizes patterns and classifies data, by producing creative outputs.

The core technology of generative AI is learning the fundamental patterns and structures of data based on large datasets [14]. For example, LLMs learn vast amounts of text data to understand grammar, vocabulary, context, and even specific styles, enabling them to generate new sentences, answer questions, or summarize text. These capabilities hold potential for various applications in corporate environments, such as customer service chatbots, automated content marketing material generation, and support for software code development. As a result, companies are seeking ways to automate diverse work processes—such as code writing, RAG-based search, and image processing—using AI, with increasing demand to enhance productivity and efficiency [15].

Additionally, by utilizing LLM, it is possible to address the limitations of systems that rely solely on OCR for document digitization, such as the difficulty OCR faces in misrecognizing characters [16].

Accordingly, in complex tasks such as financial expense processing automation, which involves intricate rules and exceptions, generative AI can play a role in understanding user intent through natural language-based question-and-answer interactions, comprehending policy documents, and providing appropriate judgments. For instance, when account classification for a new item is required, generative AI can recommend the most suitable account based on existing data and policies or assist in determining whether the item aligns with company policies. This presents new possibilities for solving non-standard and context-dependent problems that were challenging for traditional rule-based automation systems.

When examining technological trends, generative AI is evolving to produce increasingly sophisticated and human-like outputs as the scale of models grows and the quantity and quality of training data improve [13]. Additionally, research on small-scale models (Small Language Models, SLMs) tailored to specific domains or tasks is actively underway, enabling the development of AI solutions optimized for particular industries or corporate environments. Furthermore, just as humans acquire and understand information through various senses such as sight, hearing, and touch, multimodal AI systems can process and comprehend multiple forms of data—such as text, images, audio, and video—simultaneously [17]. With the advancement of multimodal AI, methodologies for implementing multimodal LLM-based Multi-Agent Systems (MAS) and effectively integrating AI technologies into business processes are being proposed. However, the use of generative AI comes with ethical and technical challenges, including issues related to the factuality of generated content, bias, copyright, and potential misuse. Therefore, a cautious approach and continuous research are necessary to address these concerns [13].

## 2.2. Document Understanding Technology

Document Understanding is a technology that automatically extracts, classifies, and verifies text and data from various types of documents (paper, images, PDFs, etc.), converts them into digital data, and automates business processes. OCR and IDP play a key role in this process. Recently, it has been categorized into AI OCR, which adds AI capabilities to traditional OCR, and IDP, which enables intelligent document processing, as shown in Figure 1. IDP and OCR are complementary to each other. OCR provides the technical foundation for extracting text from document images, while IDP builds a comprehensive system that understands the meaning of documents based on OCR results, extracts necessary information, and utilizes it in business processes. Therefore, OCR plays a critical role in the data collection stage of IDP, and IDP implements the automation and intelligence of document processing by leveraging various AI technologies, including OCR.

**Figure 1**
**Document Understanding Concept Diagram**

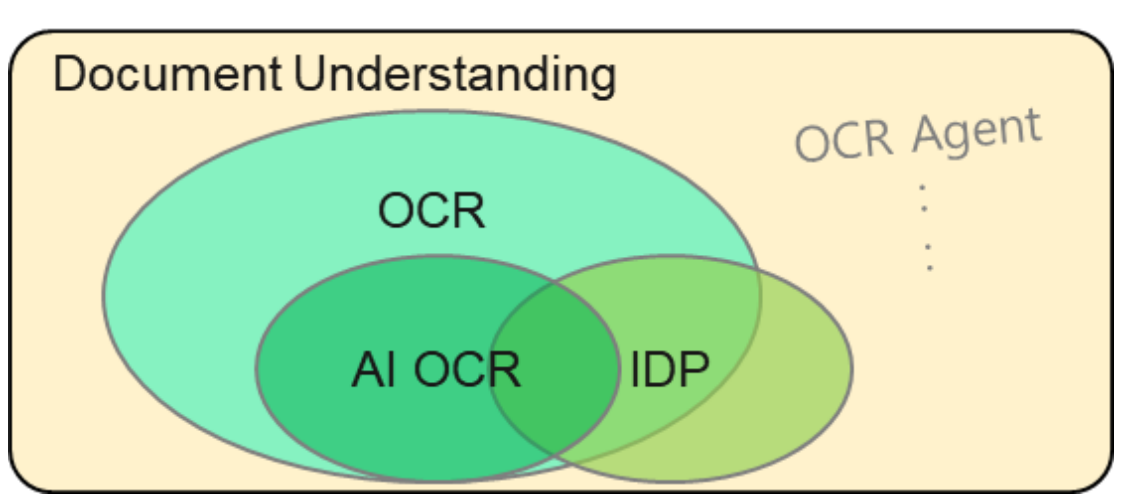

### *2.2.1. Optical Character Recognition (OCR)*

OCR is a method for extracting text from scanned documents and images [18]. CR is an optical character recognition technology that first detects character areas in images or documents through the character detection stage, and then performs the character recognition stage for those areas [19]. Generally, the OCR process is divided into image preprocessing, text segmentation/localization, feature extraction, text recognition, and post-processing [20]. Through this, OCR technology can automatically extract text from images or scanned documents and convert it into editable digital data. Early OCR technologies primarily demonstrated high recognition rates for standardized fonts and clean background documents. However, due to factors such as diverse fonts, handwriting, complex layouts, and degraded image quality, their application in real-world work environments was limited. Nevertheless, with the recent advancements in artificial intelligence, particularly deep learning-based image recognition technologies, the accuracy and scope of OCR have significantly improved [19]. The latest OCR engines support various languages and characters, providing relatively high-level text extraction performance even in low-resolution images or distorted documents.

### 2.2.2. Intelligent Document Processing (IDP)

IDP is a concept that takes OCR technology to the next level. Unlike traditional OCR, which only extracts text, IDP goes beyond by automatically classifying documents (e.g., invoices, contracts, receipts), identifying specific fields within the document (e.g., vendor name, amount, date, items), extracting them as structured data, and validating the accuracy of the extracted information. This enables more intelligent document understanding and processing capabilities.

IDP solutions utilize a combination of various AI technologies, such as machine learning (ML), natural language processing (NLP), and computer vision, to accurately extract key information from unstructured or semi-structured documents and integrate it with enterprise systems (ERP, CRM, etc.) to automate subsequent business processes [8]. In particular, understanding document images [9] is a critical yet challenging task that requires complex capabilities such as text reading and comprehensive document comprehension [21]. However, recently, generative AI technologies have been integrated into IDP systems, where LLMs are used to understand document meanings, perform context-based data extraction and classification tasks. For example, when extracting information such as purchase items, amounts, and dates from receipt images, LLMs contribute to identifying accurate information by considering the context. Additionally, LLMs are utilized in tasks such as automatically summarizing or classifying extracted data, enhancing the intelligence and efficiency of IDP systems. IDP technology continues to evolve, and recently, it has been provided in the form of cloud-based services, which reduce initial implementation costs and increase scalability. Furthermore, 'zero-shot' or 'few-shot' learning techniques have been applied, enabling learning for specific document types with only a small number of sample documents, thereby providing flexibility to quickly respond to new types of documents [1, 22]. These technological advancements are playing a crucial role in expanding the scope of automation in document-intensive fields such as finance, accounting, law, and human resources.

The main comparison of OCR and IDP, as previously reviewed, is shown in Table 1.

**Table 1**
**Comparison of OCR and IDP**

| Item | OCR | IDP |
|---|---|---|
| Function | Read the text, digitize it, and convert it into a searchable format. | Read the text, understand its meaning using AI and NLP, and classify and analyze the data. |
| Subject Matter | Structured Data (Fixed Format) | Structured, semi-structured, and unstructured data can all be processed. |
| Technical Components | Image Enhancement, Object Detection, Text Recognition (OCR/ICR) | OCR + AI (NLP, ML), Rule-Based Analysis |
| Main Use Cases | PDF Conversion, Digital Archiving, Search Function | Loan Application Processing, Customer Due Diligence (KYC), Insurance Claim Automation |
| Contextual Understanding Ability | None | Presence (able to understand context and extract additional information) |
| Automation Potential | Limited (restricted to simple text extraction) | Advanced automation capabilities (including workflow triggers and decision support) |

In financial expense management tasks, OCR/IDP technology is used to perform automation. The receipts, invoices, and transaction statements submitted by employees are highly diverse in form and format, and are mostly composed of unstructured data. Previously, financial personnel had to manually review these documents and input the necessary information into the system. However, with the introduction of IDP, this process can now be automated. For example, when a user uploads a corporate card receipt image to the system, the IDP solution automatically recognizes and extracts information such as the company name, business registration number, transaction date, supply amount, tax amount, total amount, approval number, and individual purchase item names, quantities, and unit prices from the image. In the Automation Agent case study of this research, IDP was utilized as the core technology in the first stage of the "document recognition" process, where it extracted text such as item names, amounts, and dates from receipt images and converted them into usable data. Notably, IDP's ability to go beyond simple OCR by understanding context and accurately classifying items significantly enhances the accuracy and efficiency of subsequent automation stages.

## 2.3. RPA and Automation Agent

RPA is a technology that algorithmizes simple and repetitive tasks to automate them through software, enabling them to be performed in place of humans [23]. It is also a method of automating tasks that users perform repeatedly and simply on computers by allowing software to execute them instead [24]. In this way, RPA is a technology that automates tasks, meaning the automation of service tasks performed by humans [25]. Additionally, RPA automation complements people by performing all or part of the functions previously performed by humans to achieve work goals [26]. RPA primarily mimics user interface (UI) interactions to operate existing applications (e.g., ERP, CRM, websites, Excel), allowing for rapid automation without system changes, which is one of its advantages [1]. Initially, RPA was mainly applied to simple tasks based on clear rules and structured data, such as data entry, file transfer, email processing, and report generation. Through the introduction of RPA, companies could expect benefits such as improved work processing speed, reduced errors, cost savings, and employees' focus on high-value-added tasks [1].

However, traditional RPA had several limitations. First, it struggled with unstructured data processing. It lacked the ability to understand and process various forms of unstructured data, such as images, natural language text, and scanned documents, making it difficult to apply to tasks involving such data. Second, it was inflexible in handling exceptions. When exceptions occurred outside of predefined rules, RPA bots often halted their tasks or generated errors, frequently requiring human intervention. Third, it had limitations in complex decision-making. It was challenging to implement decision-making logic that required consideration of multiple variables or human judgment [1].

To overcome these limitations, the concept of Intelligent Automation or Hyperautomation has emerged. The main technologies applied in Hyperautomation include cognitive and execution technologies that recognize, execute, and make decisions about tasks and processes to be automated, enabling optimal automation [19]. Achieving Hyperautomation requires providing services in a more automated manner to increase speed, reduce errors, and deliver them more consistently [27]. The goal is to expand the scope of automation and infuse intelligence by combining process automation technologies such as RPA with various technologies like AI, ML, NLP, and IDP. In the field of process automation, next-generation AI-based agents capable of executing complex tasks have emerged, with Automation Agents playing a crucial role [28]. Automation Agents are no longer just simple task performers; they are intelligent software robots that can recognize more complex situations, analyze data, interact with users, and even improve their performance through learning by integrating AI technologies.

The Automation Agent serves as both an interface and an executor, enabling corporate financial managers to comprehensively review and make final decisions on information extracted by IDP, database comparison results, and Gen AI recommendations. Notably, it plays a pivotal role in implementing a Human-in-the-loop[1] mechanism by automatically reflecting financial managers' judgments (e.g., approving account classifications for new items) into the system. This allows the system to learn and make decisions autonomously in similar cases in the future. By involving humans to handle exceptions and enabling the system to learn from these outcomes, it creates a virtuous cycle that continuously enhances the intelligence and adaptability of the automation system. Ultimately, modern RPA is evolving beyond simple rule-based automation, leveraging AI-powered Automation Agents to collaborate with humans and adapt flexibly to complex and dynamic work environments. This evolution is essential for achieving E2E automation in areas such as corporate expense management, where exceptions are frequent and human judgment is critical.

## 2.4. E2E (End-to-End) Automation

RPA E2E automation refers to the process of automating an entire workflow from start to finish with minimal or no human intervention. It moves beyond fragmented approaches that only automate individual tasks or specific segments, aiming instead to integrate and optimize the overall workflow by seamlessly connecting multiple systems, departments, and diverse technological elements. Additionally, it is advisable to select processes that can be completed within 1 to 2 weeks as the target for E2E task handling, as this duration is manageable for monitoring and oversight on the platform. If the process becomes too lengthy, it may require continuous monitoring and management until completion, which could hinder rather than enhance automation efficiency [19].

---

[1] The term human-in-the-loop (HITL) generally refers to the need for human interaction, intervention, and judgment to control or change the outcome of a process [29].

E2E automation goes beyond simply replacing repetitive tasks, focusing on supporting data-driven decision-making, flexibly addressing exceptions, and ultimately maximizing business value [1].

Traditional task automation has primarily been carried out at the task level. For example, individual tasks such as data entry, report generation, and data synchronization between systems could be relatively easily automated using tools like RPA. However, in reality, most corporate business processes are complex, involving multiple stages, various systems, and numerous stakeholders. In such an environment, fragmented automation has limitations in resolving bottlenecks in the entire process or achieving significant efficiency improvements. Instead, it often led to new inefficiencies due to the disconnect between automated and manually processed parts.

E2E automation is a concept that emerged to overcome these limitations, holding significance as shown in Table 2.

**Table 2**
**Benefits of E2E Automation**

| Classification | Description |
|---|---|
| Enhanced Process Visibility and Control | Monitor and manage the entire process from the beginning to the end of operations to identify and improve bottlenecks or inefficiencies, ensuring compliance with regulations and risk management. |
| Maximizing Operational Efficiency | Minimize manual work and seamlessly connect data flow between multiple systems to reduce overall processing time and operational costs, contributing to improved productivity [1]. |
| Data-Driven Decision Support | Collect and analyze data generated throughout the process to enable decision-making based on objective indicators. |
| Enhancing Customer and Employee Experience | As the speed and accuracy of task processing improve, customer satisfaction increases, and employees can focus on more creative and strategic tasks, freeing them from repetitive and low-value work. |

The automation of the 'E2E Financial Expense Processing' process using the Automation Agent is a prime example of the importance of E2E automation. From attaching receipts to extracting information through IDP, policy-based classification, exception handling using AI Flow, and final review and system learning through the Automation Agent, the entire process of financial expense processing is seamlessly connected and automated. This goes beyond replacing repetitive tasks to effectively implement complex scenarios that require human judgment. In this way, E2E automation plays a key role in enabling companies to achieve true digital transformation and enhance their competitiveness.

## 2.5. An Examination of Corporate Financial Expense Management Tasks

The task of managing corporate financial expenses is a universal and essential administrative task that occurs in all organizations, regardless of their size or industry. It involves the process of reimbursing and paying expenses incurred by employees in relation to their work (e.g., transportation costs, meal expenses, purchase of supplies, travel expenses, etc.) in accordance with the company's regulations and procedures. The task of managing expenses has the following key characteristics.

First, handling various proof documents: The most basic step in financial expense processing is collecting and reviewing documents that prove expenditures, such as receipts, tax invoices, transaction statements, and business trip requisition forms. These proof documents vary greatly in form (paper, electronic), format, and the type and level of information they contain. In particular, receipts often consist of a mix of unstructured text and images.

Second, compliance with internal regulations and policies: Each company has its own regulations and policies related to expense processing (e.g., allowable/prohibited items, limits, account classification criteria, etc.), and all expense processing must strictly adhere to these regulations. This goes beyond mere amount verification and includes a process of determining whether the purpose and content of the expenditure align with the company's policies.

Third, frequent occurrence of exceptional situations: In the financial expense processing process, various exceptional situations can occur. For example, cases where the content of a receipt does not match the actual purchased item, where a new item does not clearly fall under the existing account classification criteria due to the introduction of a new item, or where there are typos or information omissions are common. For instance, when receipts classified as consumables include items like coffee or snacks, such issues that cannot be resolved through simple automatic classification may frequently and continuously arise.

Fourth, multi-stakeholder involvement and approval process: Multiple stakeholders, including the expense applicant, department head, financial officer, and final approver, are involved in the expense processing procedure. Each step requires review and approval, ensuring transparency and accuracy in task execution. However, this process can also become a factor that delays overall processing time.

Fifth, requirement for data accuracy and audit traceability: Financial data serves as the foundation for a company's accounting processes and tax filings, thus requiring a high level of accuracy. Additionally, all processing procedures must be clearly recorded and traceable for potential audits.

Due to these characteristics, traditional manual-based financial expense processing required significant time and effort, and was prone to human errors. To address this, some companies attempted to introduce RPA to automate repetitive tasks such as data entry and information transfer between systems. However, traditional RPA faced challenges in achieving full automation of financial expense processing due to limitations such as those outlined in Table 3 [1].

**Table 3**
**Limitations of Traditional RPA**

| Limit | Description |
|---|---|
| Limitations in the Processing of Unstructured Evidence Materials | Limitations in accurately recognizing and extracting necessary information from unstructured documents such as receipts [8]. |
| Inability to respond to exceptional situations | Rule-based RPA struggles to handle unexpected situations that are not predefined [1]. |
| Complexity in Judgment and Decision-Making | The appropriateness of expense items, classification of account subjects, and review of policy violations require human cognitive judgment, making them difficult to automate. |
| Bottlenecks caused by partial automation | When only a portion of the entire process is automated, the lack of smooth integration between the automated and manual segments can lead to the emergence of new bottlenecks or impact the overall improvement in efficiency. |

In conclusion, due to the complexity, diversity, and frequent exceptions in corporate financial expense management, it has been challenging to implement effective E2E automation using only traditional simple automation technologies. To overcome these limitations, it is essential to achieve accurate receipt recognition through OCR/IDP, utilize intelligent technologies such as generative AI for exception handling and decision support, and establish a human-AI collaboration environment through Automation Agents.

The case studied in this research presents a new approach that goes beyond the limitations of existing automation by integratively utilizing these technologies.

## 3. Materials and Methods

In this chapter, based on the theoretical background discussed earlier, we present a case study of successfully implementing E2E automation for financial expense processing tasks in an actual corporate environment by utilizing generative AI and OCR/IDP-based Automation Agents. The core contribution of this study lies in addressing the limitations of existing RPA and simple automation technologies by proposing an E2E automation framework that seamlessly integrates generative AI, OCR/IDP, and Automation Agents. This framework is designed to be applied to complex and exception-prone tasks, such as corporate expense processing. Notably, the study introduces an intelligent approach to handling unstructured data and exceptions, which are traditionally challenging for RPA, by leveraging the synergistic integration of generative AI and IDP. Furthermore, the incorporation of a Human-in-the-loop mechanism through Automation Agents ensures that the final judgment of responsible personnel is reflected within the system, enabling continuous improvement in automation performance. Additionally, the system is architected with an API-based loosely coupled structure, facilitating scalability across diverse business domains, including accounting, human resources, and procurement. This design underscores a significant technological innovation with practical applicability.

The subject of this study, Company S, is a major domestic corporation operating various business divisions and overseas branches, requiring the processing of a large number of corporate card receipts on a monthly basis. Expense management is a critical process for financial management and compliance with regulations at S, but the existing manual-based processing method was fraught with several issues. Therefore, this chapter aims to present the practical applicability and effectiveness of the proposed technology integration by detailing the problems of the existing process, the background and objectives of introducing an automation system, and the specific four-stage implementation process.

## 3.1. Technical Approach and Implementation Method

In this section, we will describe in detail the technical approach and implementation of the end-to-end automation system for financial expense processing using Automation Agent, as well as the integration plan for its key components. The financial expense processing automation based on Automation Agent proceeds in four stages, as shown in Figure 2. The discussion focuses on the system architecture supporting this four-stage automation process, the core technologies such as generative AI, OCR/IDP, and the integration mechanism of Automation Agent, and the overall flow of data.

**Figure 2**
**Business Automation Process Based on IDP and Automation Agent**

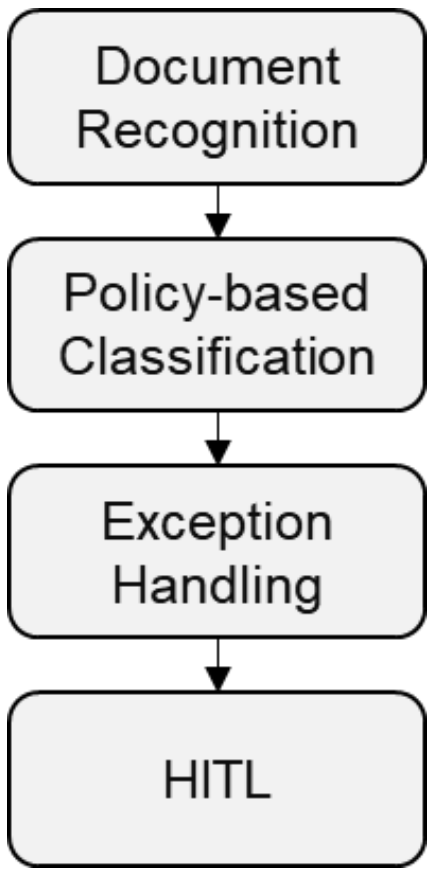

### *3.1.1. Overview of Automation Agent Application*

Today, companies are moving beyond the automation of simple repetitive tasks to pursue flexible automation that considers exceptional situations, i.e., E2E automation combined with human judgment. However, building an automated environment that can flexibly incorporate human judgment in complex and diverse exceptional situations is a challenging task. To address this issue, this study applied Brity Automation[2], an automation-focused Automation Agent, to automate the financial expense processing tasks of a company, rather than directly developing a program source code.
The main reasons for applying Brity Automation are as follows.

- **Increased demand for intelligent automation:** Beyond the limitations of simple RPA, there is a growing need for intelligent automation solutions that can implement complex end-to-end scenarios by leveraging Gen AI, IDP, and BPA (Business Process Automation).
- **Improving Work Efficiency and Maximizing Productivity:** To alleviate the workload of financial personnel and enable them to focus on more valuable tasks, we aimed to address inefficiencies in financial expense processing, which accounted for an average of 1,450 cases per month and took more than 24 hours to complete.
- **Enhancing Support for Partners and Customers:** It was necessary to create and provide use cases to help customers who had adopted RPA expand their automation areas and reduce trial and error for partners when applying new features of Brity Automation.

The goal of financial expense automation using Brity Automation has been set as follows under this background.

- **Implementing End-to-End Automation:** From receipt submission to final processing and system learning, the entire financial expense process is integrated and automated.
- **Exception Handling Intelligence:** Implement intelligent judgment support for new items and ambiguous cases through accurate information extraction using IDP and AI Flow[3] (LLM integration).
- **Establishing a Human-AI Collaboration Environment:** Through the Automation Agent, finance professionals review AI suggestions and make final decisions, and the system learns from these outcomes to continuously improve its automation performance, establishing a Human-in-the-Loop mechanism.

[2] A hyper-automation solution for enterprise process automation utilizing AI technologies launched by SAMSUNG SDS [30].
[3] Brity Automation's Langflow-based AI workflow builder lets you easily connect AI components like LLMs, data sources, and prompts to design complex applications.

- **Flexibility and Scalability:** Establish a foundation to respond flexibly to various exceptional situations and to expand the automation model to other business areas such as accounting, human resources, and procurement in the future.

In this paper, through the case of automating the "financial expense processing" business process, which is routine yet frequently involves exceptional situations, using Brity Automation, we aim to explore the extent to which the scope of automation can be expanded. Through this case study, we suggest that it holds significant importance in exploring the potential of intelligent automation.

### *3.1.2 Integration of Main Technologies Used*

The core of this system lies in the effective integration of three major technologies: OCR/IDP, AI Flow leveraging generative AI, and Automation Agent. The integration approach for each technology is as follows.

- **Integration of OCR/IDP with Workflow Engine:** When a user uploads a receipt, the workflow engine sends the image to the OCR/IDP module. The IDP module processes the image and returns the extracted structured data in JSON or XML format to the workflow engine. This data is used as input for subsequent processing steps (policy-based classification, AI Flow invocation, etc.). Brity Automation can integrate external IDP solutions via API calls or utilize the integrated IDP functionality within the platform.
- **Policy-Based Classification Engine and AI Flow Integration with Gen AI:** When item information extracted through IDP is not clearly classified by the policy-based classification engine (not found in the whitelist/blacklist) or requires additional judgment, the workflow engine calls the AI Flow module. AI Flow sends the item information along with a predefined prompt (e.g., "Which account should this item be classified under according to the company's expense processing policy, and is it permissible? What is the basis for this?") to the generative AI model (LLM) API. The LLM generates a response based on its learned knowledge and provided company policy documents, which is then returned to the AI Flow module. During this process, the LLM may also utilize web search functionality to reference up-to-date information or external context.
- **Integration of AI Flow (LLM Interface) and Automation Agent:** Recommendations generated by AI Flow (LLM) regarding account compliance, policy alignment, and justification are presented to the finance team through the Automation Agent interface, along with other information (IDP extraction results, policy DB comparison results). Based on this information, the finance team makes the final decision, and if necessary, they can modify the recommendations or input additional details.
- **Integration of Automation Agent with Backend Systems (DB, Orchestrator):** The final decisions made by the finance team through the Automation Agent (approval, rejection, modified information, new learning data, etc.) are recorded and updated in the relevant databases (policy DB, processing history DB, etc.) via the orchestrator. If classification criteria for new items are finalized, they are reflected as whitelist/blacklist or AI model learning data, enhancing the system's intelligence. Additionally, the final processing results are transmitted to ERP or accounting systems via APIs, enabling automatic subsequent accounting processes.

Such technological integration is implemented in a loosely coupled manner based on APIs, enabling independent upgrades or replacements of individual technical components and enhancing the overall flexibility of the system.

### *3.1.3 Data collection, processing, analysis, and reflection flow*

The data flow in the E2E automation system for financial expense processing follows a cyclical lifecycle as follows.

1. Data Collection:
    - When a user submits an expense request, they input related information (such as purpose of use, amount, etc.) along with the receipt image (unstructured data) into the system.
    - The system collects uploaded receipt images as the primary input data.
2. Data Processing:
    - **1-Stage Process (IDP):** The collected receipt images are sent to the OCR/IDP module, where they undergo text extraction and key field identification processes to be converted into structured data (item name, amount, date, etc.).
    - **2-Stage Process (Policy-Based Classification):** The transformed structured data is compared with the database based on the policies of the predefined financial accounts set within the Brity Automation Orchestrator, and is initially automatically classified according to predefined rules (e.g., allow/block, account recommendations, etc.).
3. Data Analysis & Decision Support:
    - **3-Stage Process (AI Flow):** When exceptions (unclassified or ambiguous items) occur during the first classification, the data is forwarded to the AI Flow module. AI Flow utilizes Gen AI (LLM) to comprehensively analyze company policy documents, past processing cases, and, if necessary, external web

information to generate recommendations for the account, policy compliance status, and the basis for judgment.

4. User Review & Final Decision
    - **4-Stage (Automation Agent):** The extraction results from the IDP, policy-based classification results, and analysis and recommendations from AI Flow are integratedly provided to the finance team through the Automation Agent interface. The finance team reviews these and makes a final decision to approve, reject, or modify.
5. Data Reflection & System Learning:
    - The final decision of the finance officer is immediately reflected in the system. Approved expenses are transferred to the subsequent payment processing procedure, and related information can be integrated into the accounting system.
    - Especially important is that the results of the responsible person's judgment (e.g., classification of new items, updated information) are fed back and stored as learning data for the policy database or AI model within the Brity Automation Orchestrator. Through this, the system continuously learns, improves its ability to automatically handle exceptions over time, and evolves in a direction that minimizes human intervention (Self-learning & Improvement).

This cyclical flow of data goes beyond simple one-time automation, serving as the core mechanism for implementing a learning-based automation system that continuously accumulates intelligence and improves performance. This maximizes the long-term effectiveness of E2E automation and provides a foundation for adapting flexibly to changing business environments.

### *3.1.4 E2E Proposal System Architecture*

The financial expense processing E2E automation system proposed in this study is based on the Brity Automation platform and is composed of multiple modules and technical elements integrated organically. The overall system architecture can be divided into four main layers: the data input layer, the intelligent processing layer, the user interaction and learning layer, and the backend infrastructure layer, as shown in Figure 3.

**Figure 3**
**E2E Automation System Architecture**

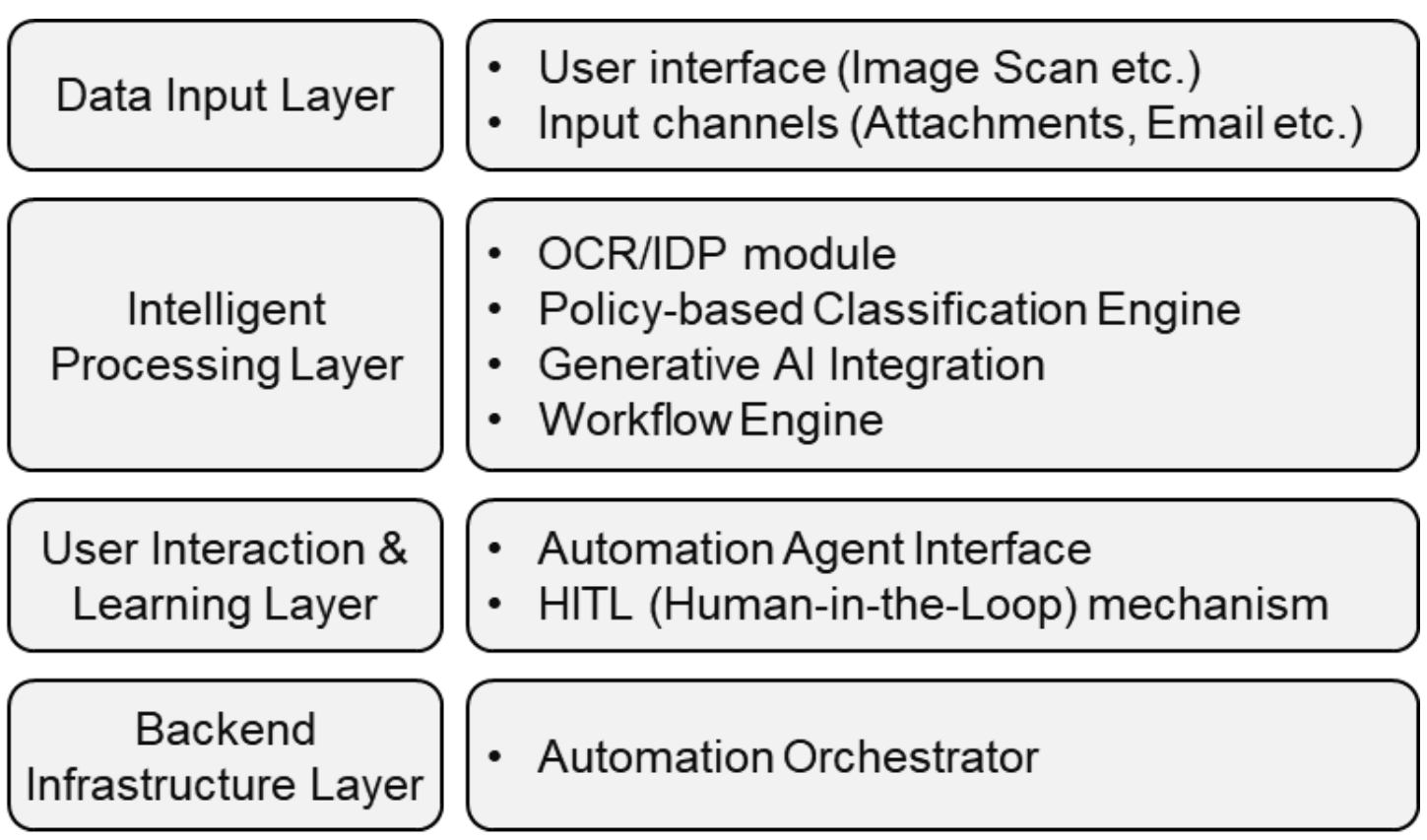

1. Data Input Layer:
    - User Interface: A web or mobile interface is provided for employees to upload receipt images (scanned or photographed) along with expense reports. This can be implemented through integration with Brity Automation's user portal or an existing groupware system.
    - Support for Multiple Input Channels: Flexibility to collect evidence materials through various channels such as email attachments, direct capture via mobile apps, and more.
2. Intelligent Processing Layer (IPL):
    - **OCR/IDP Module:** Extracts text from uploaded receipt images and identifies key fields (business name, amount, date, items, etc.) to convert them into structured data. It integrates external or open-source IDP solutions in library form and enhances the accuracy of item classification through contextual understanding.
    - **Policy-based Classification Engine:** Built within the Brity Automation Orchestrator, it performs initial automatic classification based on item information extracted through IDP, referencing an internal database

(whitelist/blacklist/account policy information). This database reflects the company's expense processing regulations and policies. When using LLM, this can be addressed through prompts, which are detailed and customized to describe the expert identity of each agent, including profile, objectives, and constraints [31].

- **AI Flow Module (Gen AI Integration):** When exceptions (new items, ambiguous entries, etc.) that were not handled in the first classification occur, this module integrates with generative AI (LLM) to provide intelligent judgment. It utilizes pre-trained company policy documents and external web search results to query the LLM, deriving recommended accounts, policy compliance, and judgment basis. Users can also obtain additional information or request clarification through a natural language interface with the LLM.
- **Workflow Engine:** A core function of the Brity Automation Orchestrator, it defines the entire financial expense processing process from data input to final handling, and automates task flows, conditional branching, notifications, and more at each stage.

3. User Interaction & Learning Layer:
    - **Automation Agent Interface:** Provides an integrated dashboard that allows financial officers to comprehensively review AI analysis results (IDP-extracted information, policy-based classification results, AI Flow recommendations) and make final decisions. This interface must be intuitive and user-friendly, designed to present all necessary information at a glance.
    - **Human-in-the-Loop (HITL) Mechanism:** Decisions made by finance personnel through the Automation Agent (e.g., confirming accounts for new items, adding similar terms) are not limited to one-time processing. Instead, they are automatically fed back and stored in the system's knowledge base (policy DB, AI model, etc.). Through this, the system continuously learns and becomes capable of making more accurate and autonomous judgments in the future when similar cases arise.
4. Backend Infrastructure Layer:
    - **Brity Automation Orchestrator:** Acts as the central control tower that manages, monitors, and executes the entire automation process. It performs functions such as workflow management, bot scheduling, database integration, API integration, security, and access control.
    - **Database:** Stores and manages policy information (whitelist/blacklist), processing history, learning data, user information, etc. In this case, the Brity Automation self-orchestrator database is used to ensure efficient maintenance and security.
    - **API Gateway:** Provides an interface for safe and efficient data integration with external systems (ERP, accounting systems, groupware, etc.).

This hierarchical architecture enables organic integration while maintaining the independence of each module, contributing to enhancing the system's flexibility, scalability, and maintainability.

## 3.2 Business Process Analysis and Design

Expense management is a representative administrative task experienced by nearly all office workers, involving the process of submitting a receipt after using a corporate card, followed by approval from the approver and the finance team, and finally being processed.

### *3.2.1 Analysis of Existing Business Processes*

The entire workflow involves complex procedures such as validating the validity of expense details, verifying account codes, and checking prohibited items based on company policies. In particular, financial expense processing faced the following issues and limitations.

First, high reliance on manual work and time consumption: The most time-consuming task was the "receipt review" process. The finance team had to manually verify various types of receipts submitted and manually input or validate the necessary information into the system. In the case of Company S, reviewing approximately 1,450 receipts per month took an average of over 24 hours, which required significant human resources and time investment.

Second, frequent occurrence of exception situations and difficulty in handling them: In the financial expense processing process, various exception situations occurred frequently. For example, an employee reported purchasing batteries under the consumables account, but the actual receipt included a box of coffee, requiring reclassification to the catering account. Additionally, when new items were introduced or there were typos on receipts, data consistency decreased, necessitating additional verification processes. These exception situations were difficult to resolve with simple automated classification and required direct intervention and judgment from the finance team.

Third, the difficulty in ensuring data accuracy: Since the process relied heavily on manual work, there was always a possibility of data entry errors or omissions. Additionally, discrepancies between the user's intentions on the receipts and the actual purchase records occurred frequently, making it challenging to ensure data accuracy. This could have a negative impact on subsequent accounting processes and audit responses.

Fourth, the limitations of existing RPA: It was difficult to flexibly respond to the complexity and exceptions with the existing RPA solutions, which were mainly applied to simple repetitive tasks. To improve efficiency, we concluded that automation beyond simple RPA, which can reflect human judgment, is necessary.

These issues increased the workload of the finance department and acted as a factor hindering overall corporate operational efficiency. Therefore, there was an urgent need to introduce a more intelligent E2E automation solution that could flexibly handle exceptions and effectively combine human judgment.

### 3.2.2 E2E Automation Process Definition

The analysis of the existing work revealed that the E2E automation process to be introduced can be defined in the flow shown in Figure 4.

**Figure 4**
**Conceptual Diagram of E2E Automation Process**

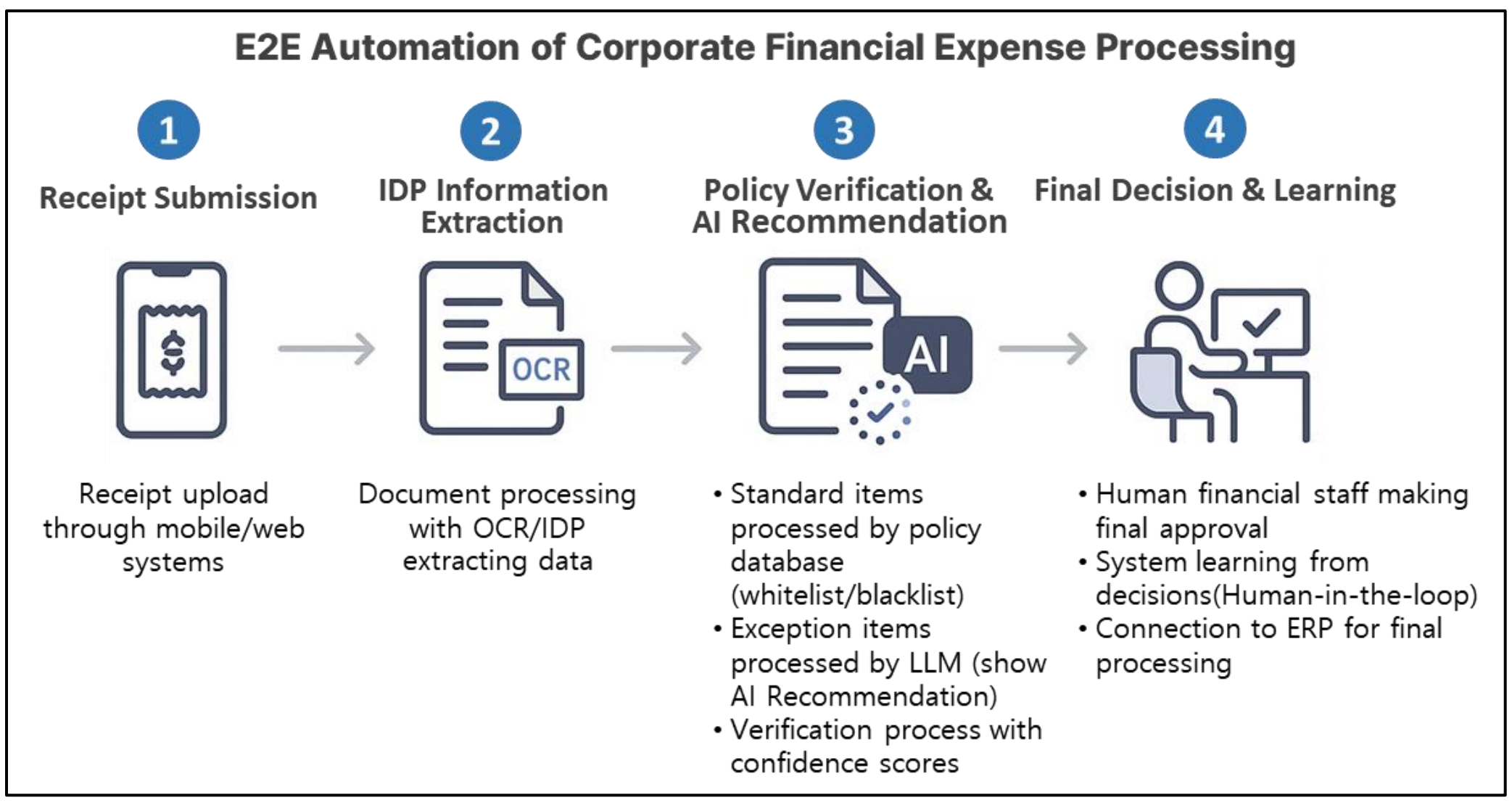

When a user registers an expense receipt, the IDP automatically extracts the necessary information from the receipt. Subsequently, the generative AI analyzes and verifies the expense details according to the company's policies, recommends the required items for system registration, and submits them. Finally, after the finance manager reviews and approves the submitted content, it is automatically registered in the ERP system.

### 3.2.3 E2E Automation Process Design

The process begins when an individual card user registers a card usage receipt through the mobile app or the card expense management screen. When processing personal expenses, the attached files are compared and verified against the input data and receipt content using RPA, IDP, and LLM to automate the manual verification process of the accounting department's account-specific responsible personnel. The entire workflow is as shown in Figure 5.

**Figure 5**
**E2E Expense Processing Workflow Design**

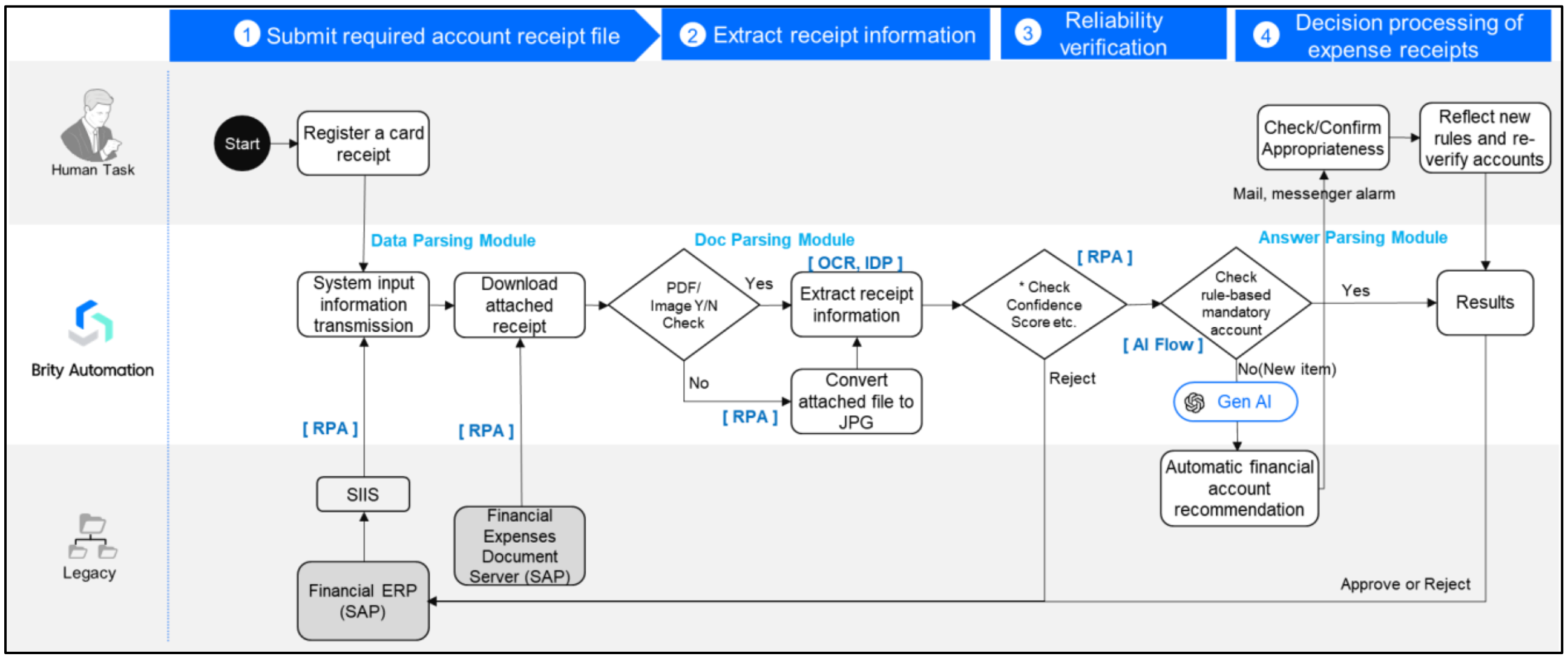

The entire processing process consists of four stages, as detailed below.

① Mandatory Account Receipt File Transmission: The card user registers the usage receipt, and the expense processing system sends the approved data (description, account, attachment) to Brity Automation.

② IDP Receipt Information Extraction: Brity Automation extracts information from the receipt attached to the received approval data. At this time, file formats other than PDF and images are converted to JPG.

③ Reliability Verification: If the Confidence score of the extracted mandatory items is below 50, the receipt is considered defective, and a notification is sent to the user. For payment data where the selected account is a routine expense, it is determined whether the detailed items extracted by IDP are appropriate for the description and account using LLM.

④ Approval Decision: If the final results, account details, and attachments are appropriate, they are approved and sent to SAP. If not, they are rejected and sent to SAP. Additionally, based on the results processed by Brity Automation, the SAP approval process is automatically progressed (Approval or Rejection).

## 4. Experimental

The financial expense automation based on the Automation Agent was implemented in four stages as shown in Figure 6.

**Figure 6**
**Business Automation Process Based on IDP and Automation Agent**

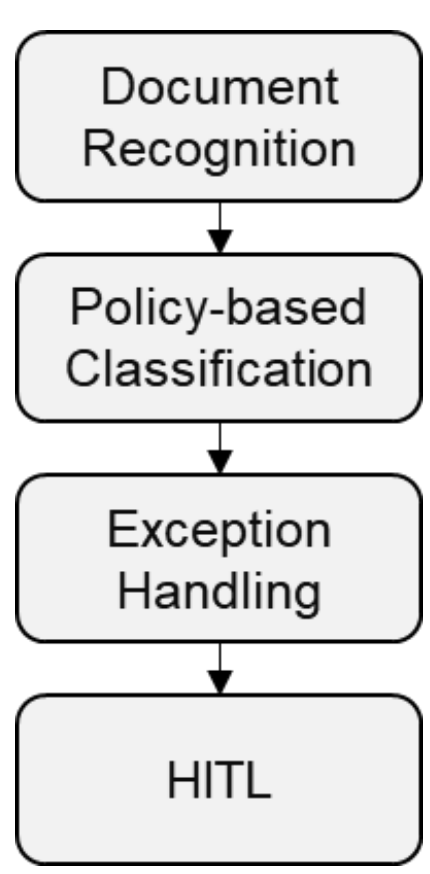

Additionally, as shown in Figure 7 implemented in the actual Automation Agent design tool, the system receives user input through the Form UI, recognizes documents using the IDP/OCR module ('T_IDP'), and then delivers the results through the Form UI after processing through the policy-based classification and exception application ('T_RULE') process.

**Figure 7**
**Integration of each module with the Automation Agent**

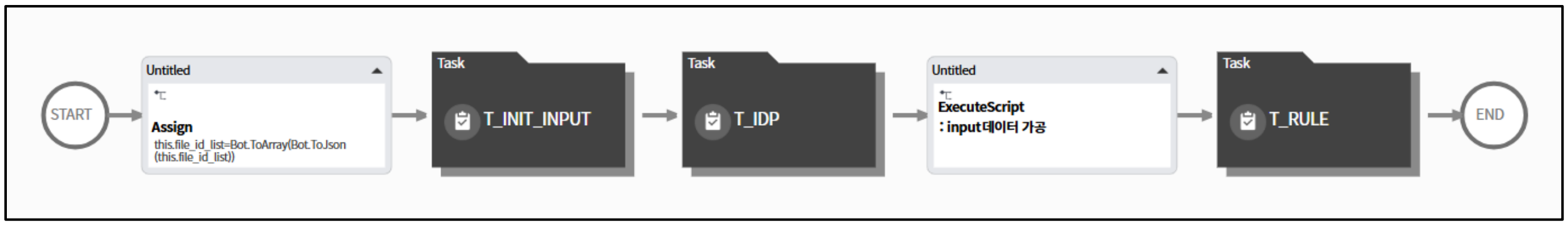

## 4.1. Step 1: Document Recognition

The first step in automation is to accurately extract the required information from the receipt image uploaded by the user using the IDP Module. To achieve this, Brity Automation leverages IDP technology.
When a user submits their card usage records, they can attach a receipt image. On a mobile app, the receipt can be automatically captured and attached, or on a PC screen, as shown in Figure 8, a receipt image can be attached. At this time, the service can choose between OCR or IDP modules, and the document type can utilize pre-trained receipt types to check or modify data extracted in advance.

**Figure 8**
**Receipt Attachment Screen**

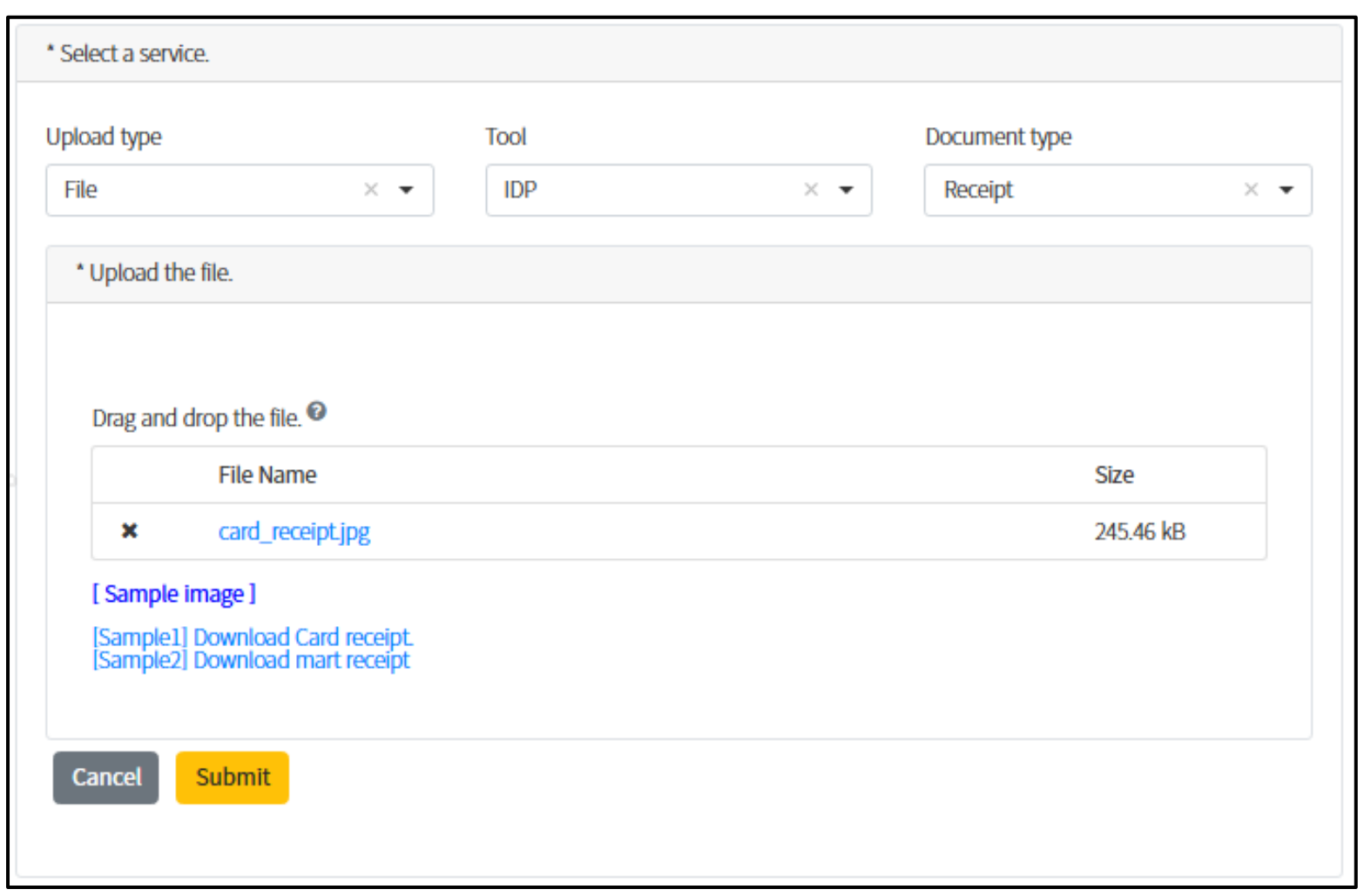

When using a corporate card, notifications are delivered to the user through a messenger app as shown in Figure 9. The image displays the usage receipt, and the notification message instructs the user to submit an expense reimbursement approval for the card usage. Upon reviewing the content, it can be confirmed that a payment of 9,000 won was made at '팝스토어잠실향군타워점(Popstore Jamsil Hyanggun Tower Branch)' on March 25, 2025.

**Figure 9**
**Receipt and Corporate Card Usage Notification**

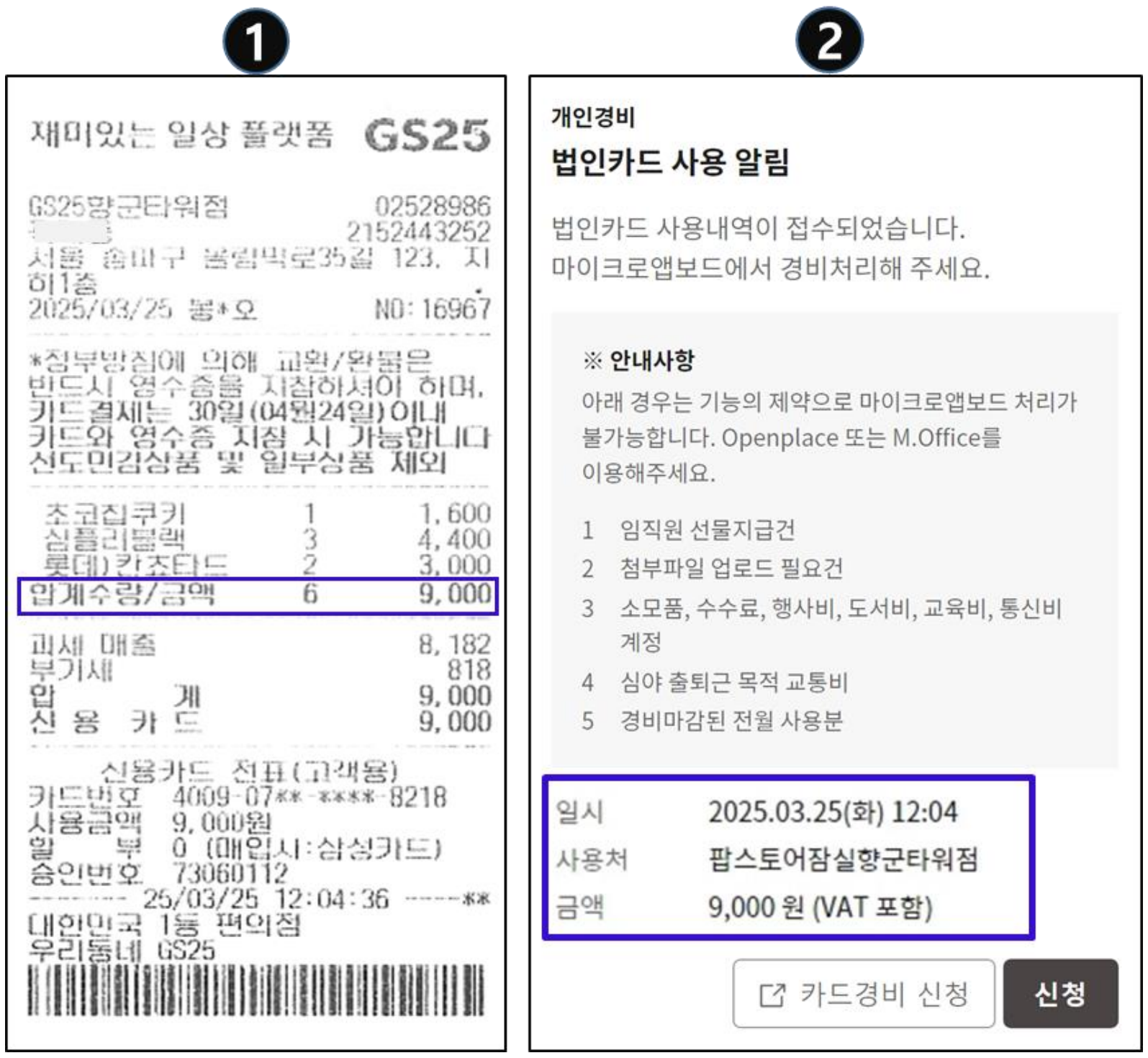

At this point, the IDP solution analyzes the image and extracts key text information such as item name, amount, and transaction date. Unlike traditional OCR technology, using IDP technology allows for understanding the context and accurately classifying items, demonstrating the intelligent information processing capabilities of IDP beyond simple text extraction. The extracted data is transformed into a user-friendly format (such as HTML or Markdown) as shown in Figure 10 and utilized in subsequent processing steps. Additionally, this process is implemented in Brity Automation to easily library external IDP using a Drag & Drop method, enhancing user convenience.

**Figure 10**
**IDP Module Design by Brity Automation**

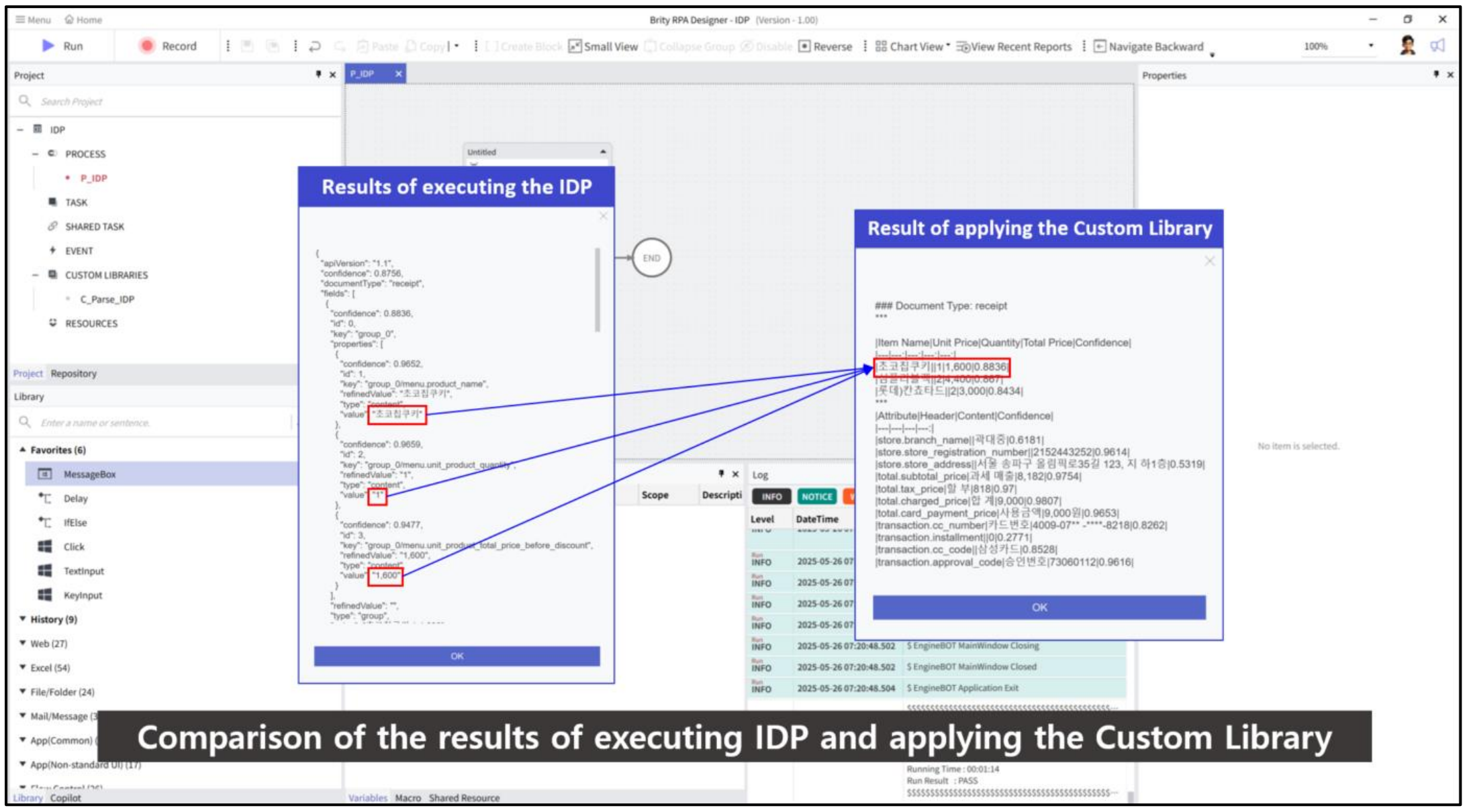

## 4.2. Step 2: Policy-based classification

When data is successfully extracted from a receipt through IDP, the task involves using Brity Automation's 'Data Service/Database' feature to determine whether the extracted item is allowed or prohibited according to the company's policy. To achieve this, Brity Automation builds its own database within the orchestrator, as shown in Figure 11, and pre-sets the types of authorized goods and product information for each account in an allowed item list (whitelist) and prohibited items in a list (blacklist).

For example, the whitelist may include general food and beverage items, while the blacklist may include inappropriate items with low relevance to work, such as gold rings or gift certificates. Additionally, depending on the company's policy, accounts such as office supplies may only be allowed for consumables.

**Figure 11**
**Policy-Based Database Configuration**

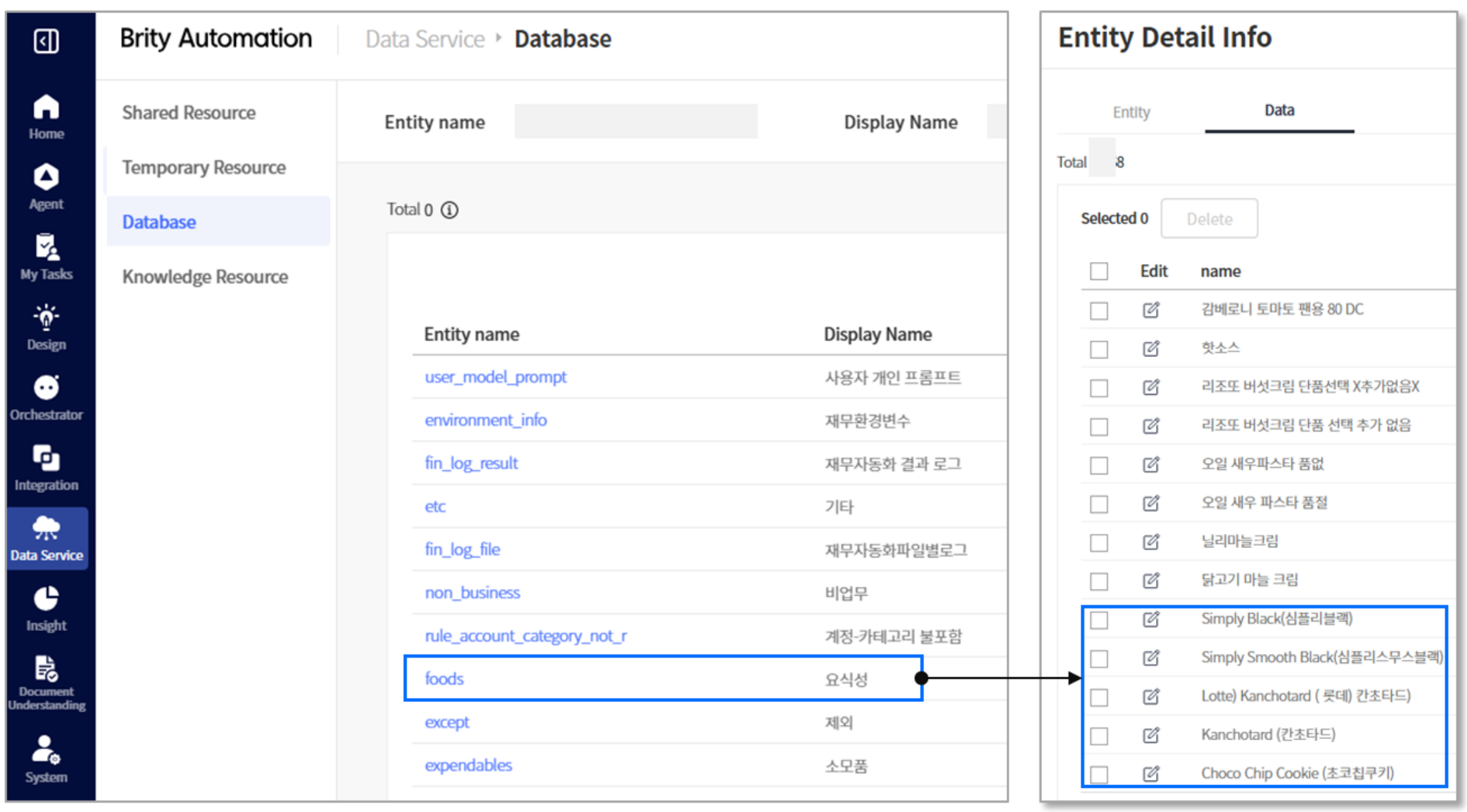

The system performs automatic filtering by comparing the extracted receipt data with the database as shown in Figure 12. This process is stored in Brity Automation's own database, not in an external one, providing efficient maintenance and security.

**Figure 12**
**Database Utilization Linked Design in Brity Automation Designer**

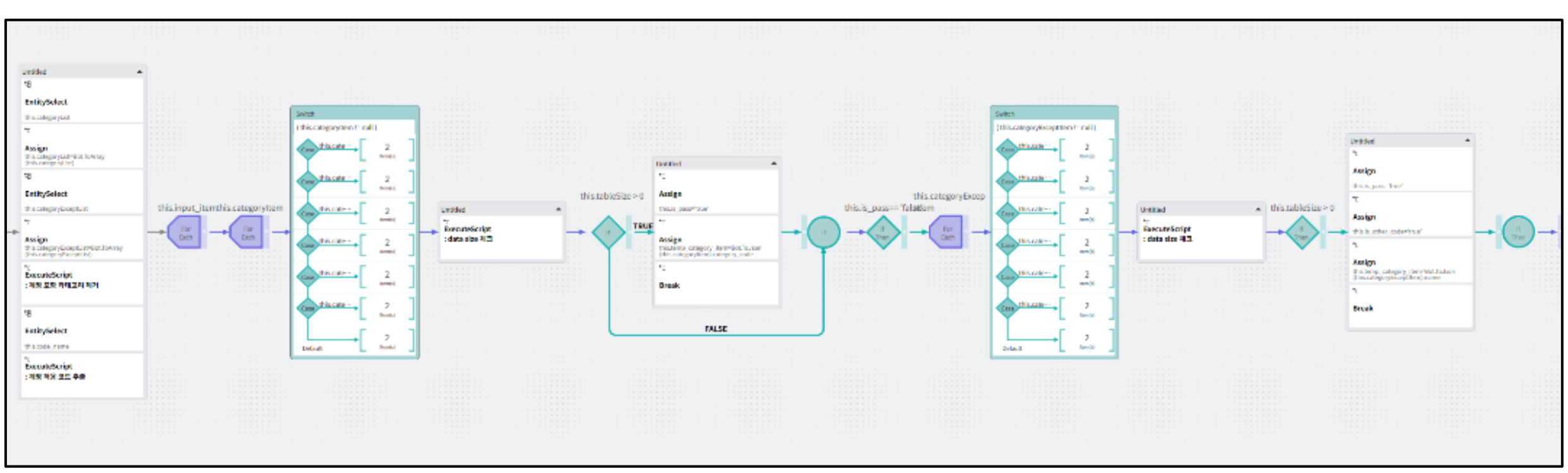

## 4.3. Step 3: Handling Exceptional Situations

Recognizing documents through IDP and initial classification based on policies alone is insufficient to address all situations. In particular, exceptions occur in cases of new items not registered in the existing database or items with ambiguous classification criteria. This was confirmed through a purchase case of a beverage called '심플리블랙(Simply Black)' When the item is a new entry not listed in the whitelist or blacklist, Brity Automation's AI Flow function is activated. AI Flow, as shown in Figure 13, queries a LLM based on the company's expense policy document to determine whether the item complies with company policy and which account should be used for processing. For example, by asking, "Which account should Simply Black be processed under?" Gen AI reviews the policy document and item classification criteria, as shown in Figure 14, and provides a recommended account along with the reasoning based on the results. In this case, the AI identifies a similar item called 'Simply Smooth Black' and recommends classifying it under a catering-related account. The role of AI is to support decision-making as a reference tool, and the final decision is made by a human.

**Figure 13**
**Design of Receipt Item Classification Using AI Flow**

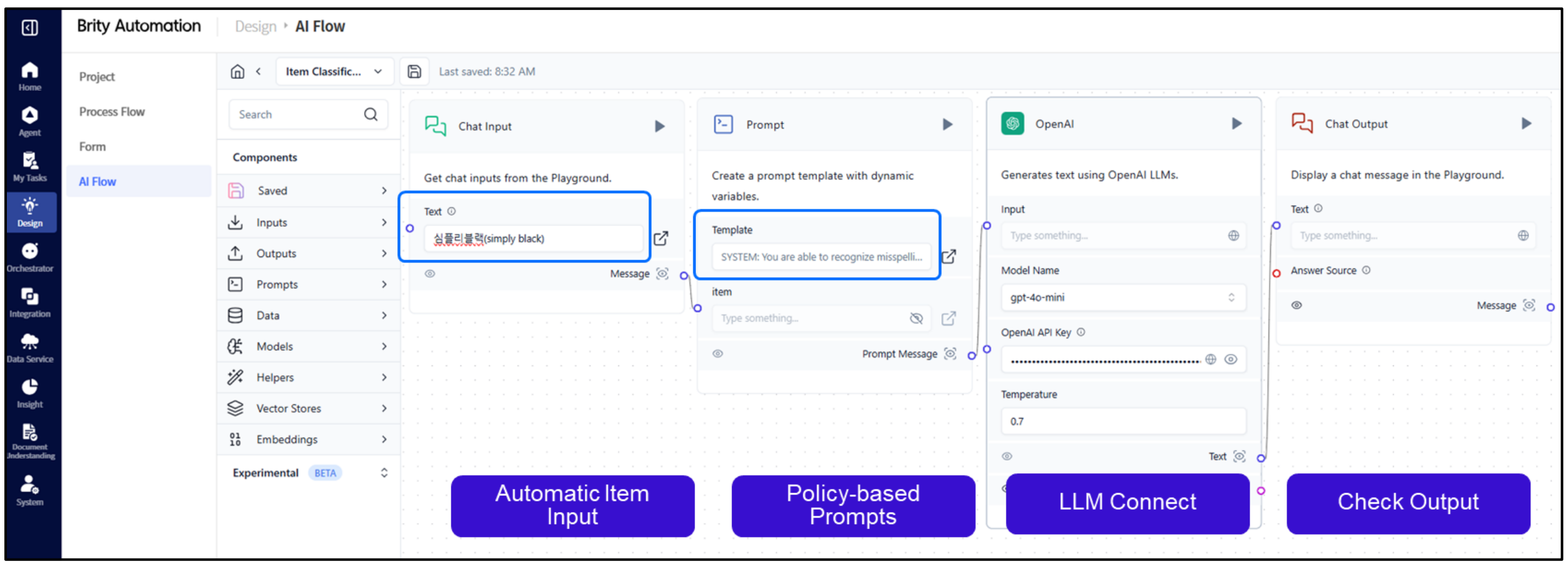

**Figure 14**
**Policy-based Prompts**

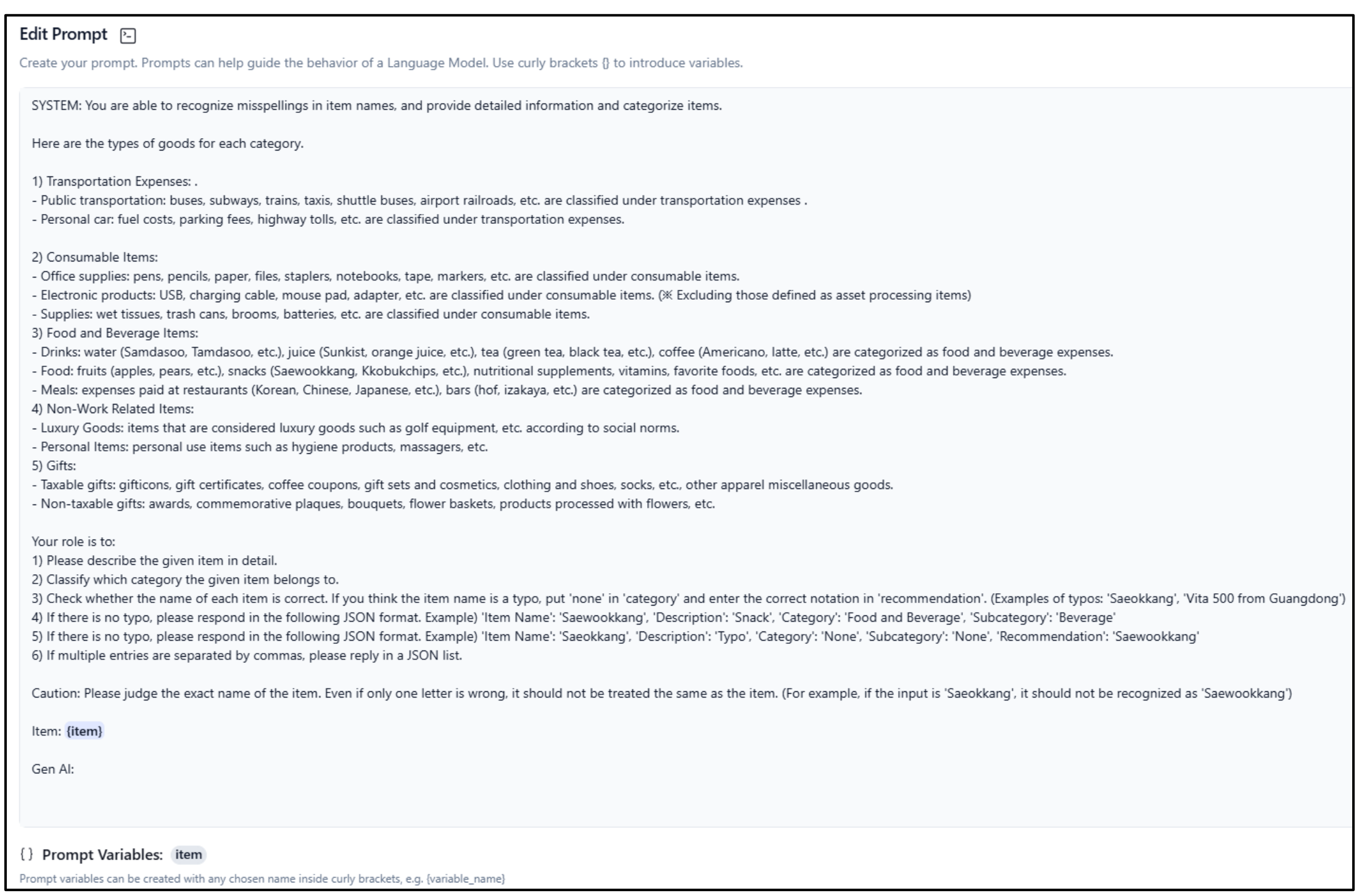

Edit Prompt

Create your prompt. Prompts can help guide the behavior of a Language Model. Use curly brackets {} to introduce variables.

SYSTEM: You are able to recognize misspellings in item names, and provide detailed information and categorize items.

Here are the types of goods for each category.

1) Transportation Expenses: .
- Public transportation: buses, subways, trains, taxis, shuttle buses, airport railroads, etc. are classified under transportation expenses .
- Personal car: fuel costs, parking fees, highway tolls, etc. are classified under transportation expenses.

2) Consumable Items:
- Office supplies: pens, pencils, paper, files, staplers, notebooks, tape, markers, etc. are classified under consumable items.
- Electronic products: USB, charging cable, mouse pad, adapter, etc. are classified under consumable items. (※ Excluding those defined as asset processing items)
- Supplies: wet tissues, trash cans, brooms, batteries, etc. are classified under consumable items.
3) Food and Beverage Items:
- Drinks: water (Samdasoo, Tamdasoo, etc.), juice (Sunkist, orange juice, etc.), tea (green tea, black tea, etc.), coffee (Americano, latte, etc.) are categorized as food and beverage expenses.
- Food: fruits (apples, pears, etc.), snacks (Saewookkang, Kkobukchips, etc.), nutritional supplements, vitamins, favorite foods, etc. are categorized as food and beverage expenses.
- Meals: expenses paid at restaurants (Korean, Chinese, Japanese, etc.), bars (hof, izakaya, etc.) are categorized as food and beverage expenses.
4) Non-Work Related Items:
- Luxury Goods: items that are considered luxury goods such as golf equipment, etc. according to social norms.
- Personal Items: personal use items such as hygiene products, massagers, etc.
5) Gifts:
- Taxable gifts: gifticons, gift certificates, coffee coupons, gift sets and cosmetics, clothing and shoes, socks, etc., other apparel miscellaneous goods.
- Non-taxable gifts: awards, commemorative plaques, bouquets, flower baskets, products processed with flowers, etc.

Your role is to:
1) Please describe the given item in detail.
2) Classify which category the given item belongs to.
3) Check whether the name of each item is correct. If you think the item name is a typo, put 'none' in 'category' and enter the correct notation in 'recommendation'. (Examples of typos: 'Saeokkang', 'Vita 500 from Guangdong')
4) If there is no typo, please respond in the following JSON format. Example) 'Item Name': 'Saewookkang', 'Description': 'Snack', 'Category': 'Food and Beverage', 'Subcategory': 'Beverage'
5) If there is no typo, please respond in the following JSON format. Example) 'Item Name': 'Saeokkang', 'Description': 'Typo', 'Category': 'None', 'Subcategory': 'None', 'Recommendation': 'Saewookkang'
6) If multiple entries are separated by commas, please reply in a JSON list.

Caution: Please judge the exact name of the item. Even if only one letter is wrong, it should not be treated the same as the item. (For example, if the input is 'Saeokkang', it should not be recognized as 'Saewookkang')

Item: {item}

Gen AI:

{} Prompt Variables: item

Prompt variables can be created with any chosen name inside curly brackets, e.g. {variable_name}

Figure 15 presents a segment of policy code that, in accordance with predefined policies, filters itemized contents from card payment receipts at the prompt level. The filtered information is then re-queried to a LLM, and the results are compared with the model's recommendations.

**Figure 15**
**Example Code for Policy Comparison Recommendation**

```
var product_list = [];
var suggestion_list = [];

// Define a redefined push function that takes value, list, and list2 as parameters.
function redefine_push(value, list, list2) {
    var product = "";

    // Check if the "recommand" property exists in the value object.
    if ("recommand" in value) {
        // Store the value of the "recommand" property in the suggestion variable.
        var suggestion = value["recommand"];
        product = suggestion;
        list2.push(suggestion);
    }
    return product;
}

// Check if conformance is an object.
if (typeof conformance === 'object') {
    // Append the length of the conformance array to the category string.
    this.category += "L:" + conformance.length + "\n";

    if (conformance.length > 0) {
        // Call the redefine_push function to process each element of the conformance array
        // and add the result to the suggestion_list.
        suggestion_list.push(redefine_push(conformance, this.redefine_list, product_list));
    }
}
```

## 4.4. Step 4: HITL (Human In The Loop) Phase

The final decision on the content proposed by AI is the responsibility of the financial officer, which is the step of automatically reflecting the user's judgment in the system. Brity Automation's messenger channel Agent provides an interface and execution environment to support the financial officer in making such final decisions effectively.

Figure 16(a) is the financial manager's messenger channel Agent screen. Through this Agent, it is possible to perform question-and-answer using RAG and execute RPA. Here, I can check and proceed with the task assigned to me, which is to confirm the card expense application agreement.

**Figure 16**

**Messenger Channel Agent Screen. (a) The Messenger home screen where an alarm is displayed when a financial manager receives a request for agreement; (b) New Item Addition Confirmation Screen;**

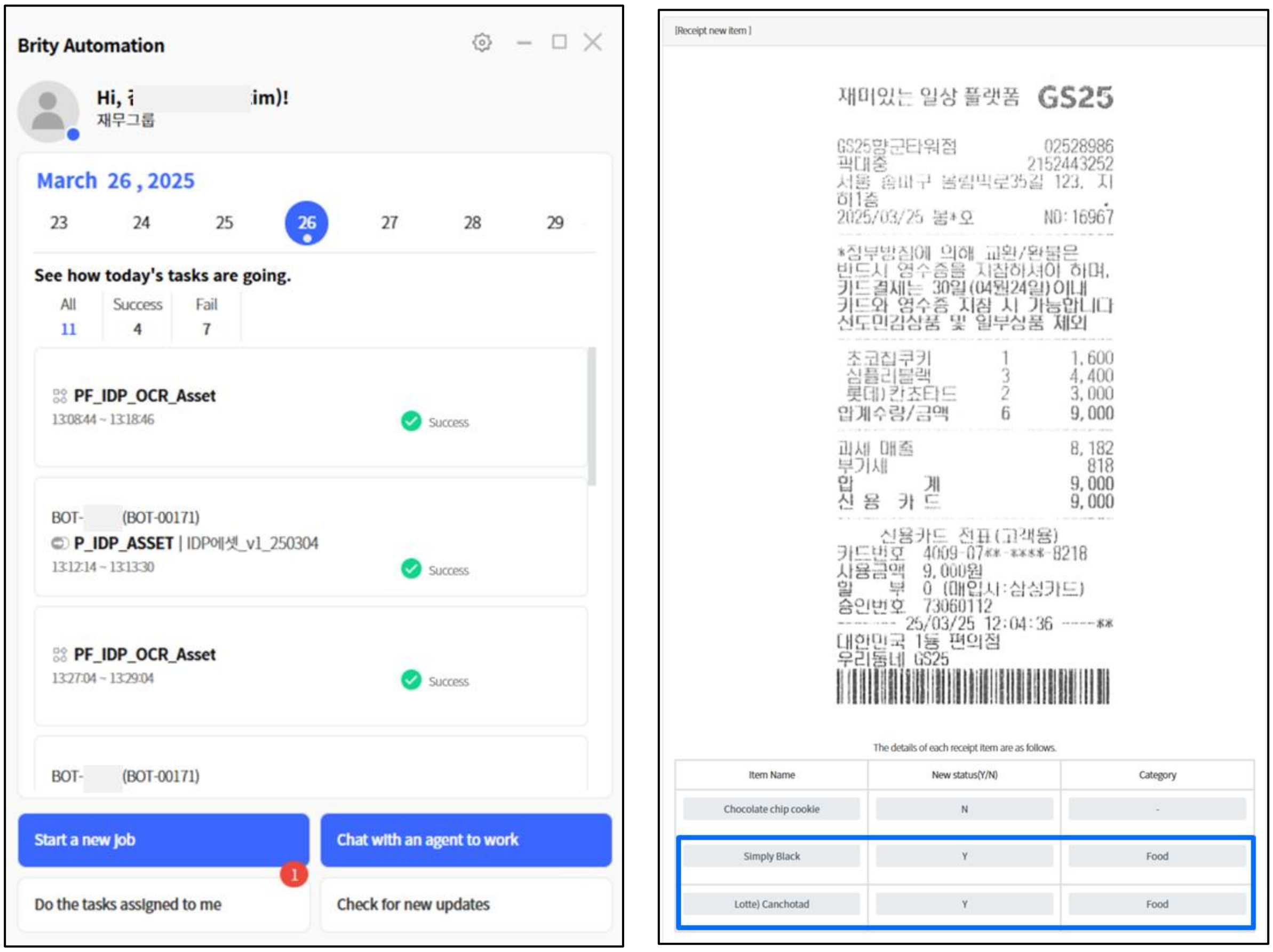

(a) (b)

The responsible party can comprehensively view the receipt image extracted by the generative AI through IDP, the comparison results with the existing database (whitelist/blacklist), and the content recommended by the AI Flow through LLM (account, policy compliance, basis, etc.) through the messenger channel agent screen using the Form UI, as shown in Figure 16(b). On this screen, the finance manager can either accept the generative AI's proposal as is or, as needed, modify the item name or add similar words to improve the system's knowledge base.

**Figure 17**
**New Item Recommendation and Data Recommendation Screen**

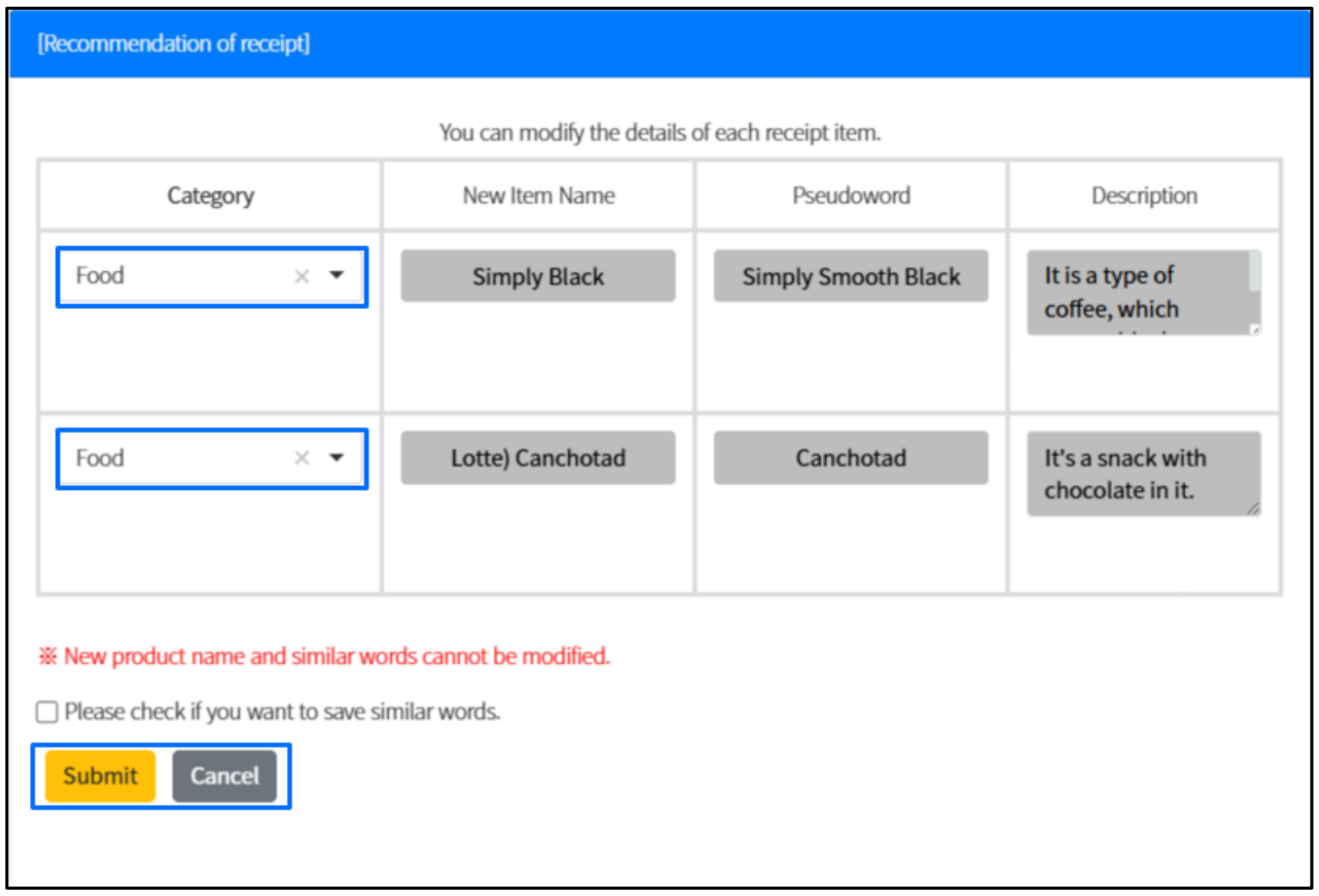

**Figure 18**
**New Item Recommendation and Data Approval Screen**

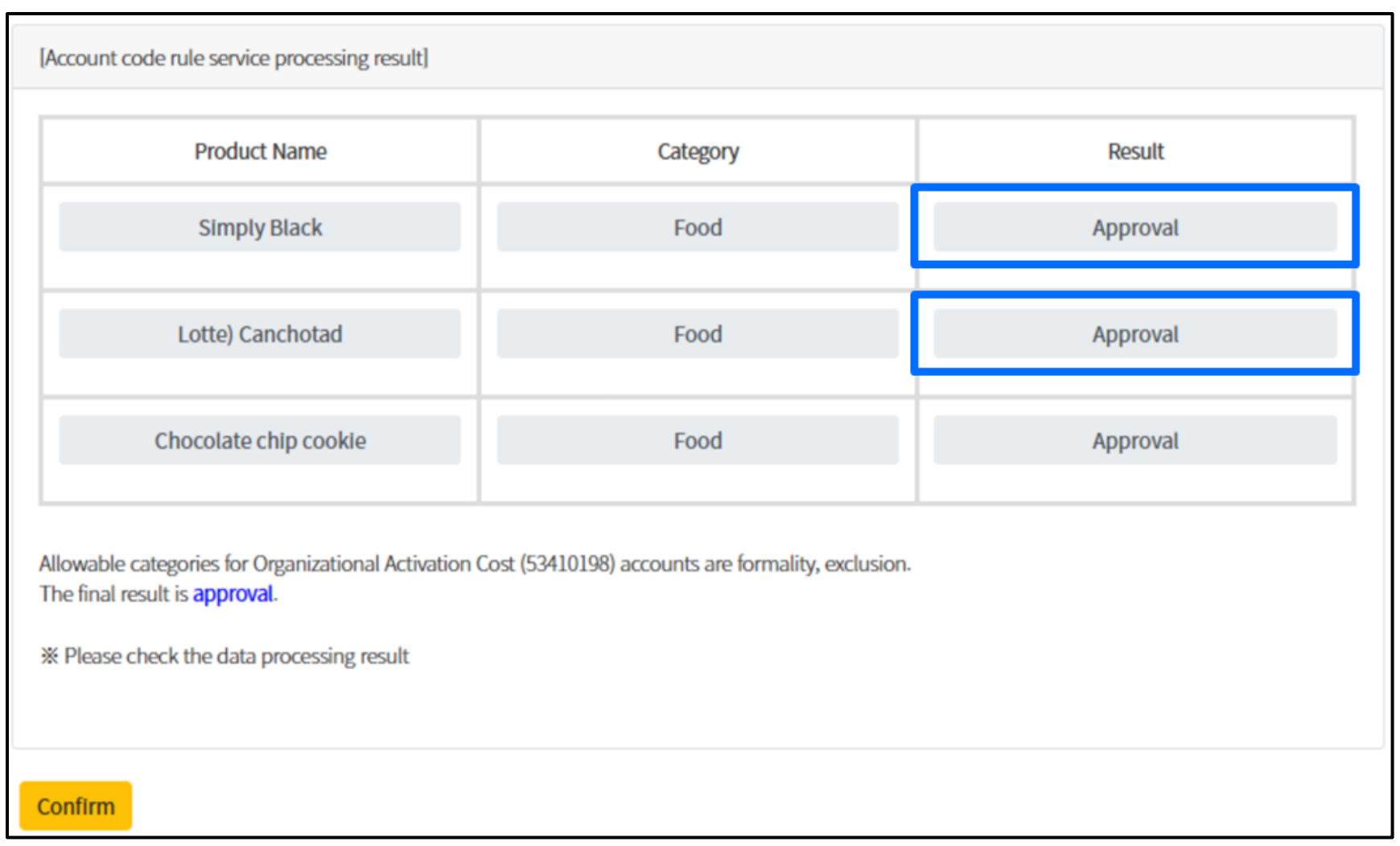

The most critical feature is that when a financial officer reviews and approves recommendation data for a new item, as shown in Figure 17, the decision does not only affect the current case but is automatically reflected in the database within the Brity Automation Orchestrator, as shown in Figure 18. This allows the system to autonomously perform the correct processing based on past learned judgments when similar items or situations occur in the future. For example, if "Simply Black" is identified as a coffee beverage and classified under a food and beverage account, this information is registered in the system, enabling automatic processing the next time the same item occurs. Additionally, by storing similar words together, the system can flexibly respond to spelling errors or subtle differences in brand names. As shown in Figure 19, by automatically incorporating user judgments, the Automation Agent effectively integrates human intellectual judgment into the system and serves as a key component of the Human-in-the-loop mechanism, gradually enhancing the intelligence of the automation system through continuous learning.

**Figure 19**
**HITL Process**

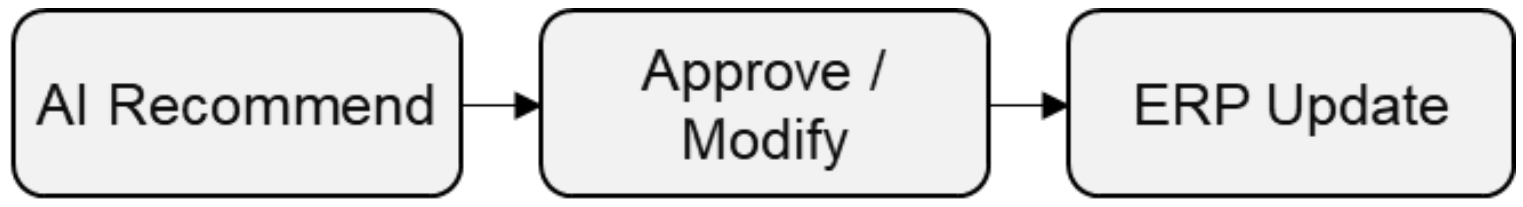

Through the above 4-stage automation process, Brity Automation completes the E2E automation of financial expense processing tasks. From the initial stage where users submit receipts, through information extraction using IDP, policy-based automatic classification, intelligent exception handling support using AI Flow, and final review by financial personnel through Automation Agent, and up to system learning, the entire Task process for each step is integrated and executed organically within the Process Flow function of the Brity Automation Orchestrator, as shown in Figure 20.

**Figure 20**
**Human Task and Overall Workflow Integration Process Flow**

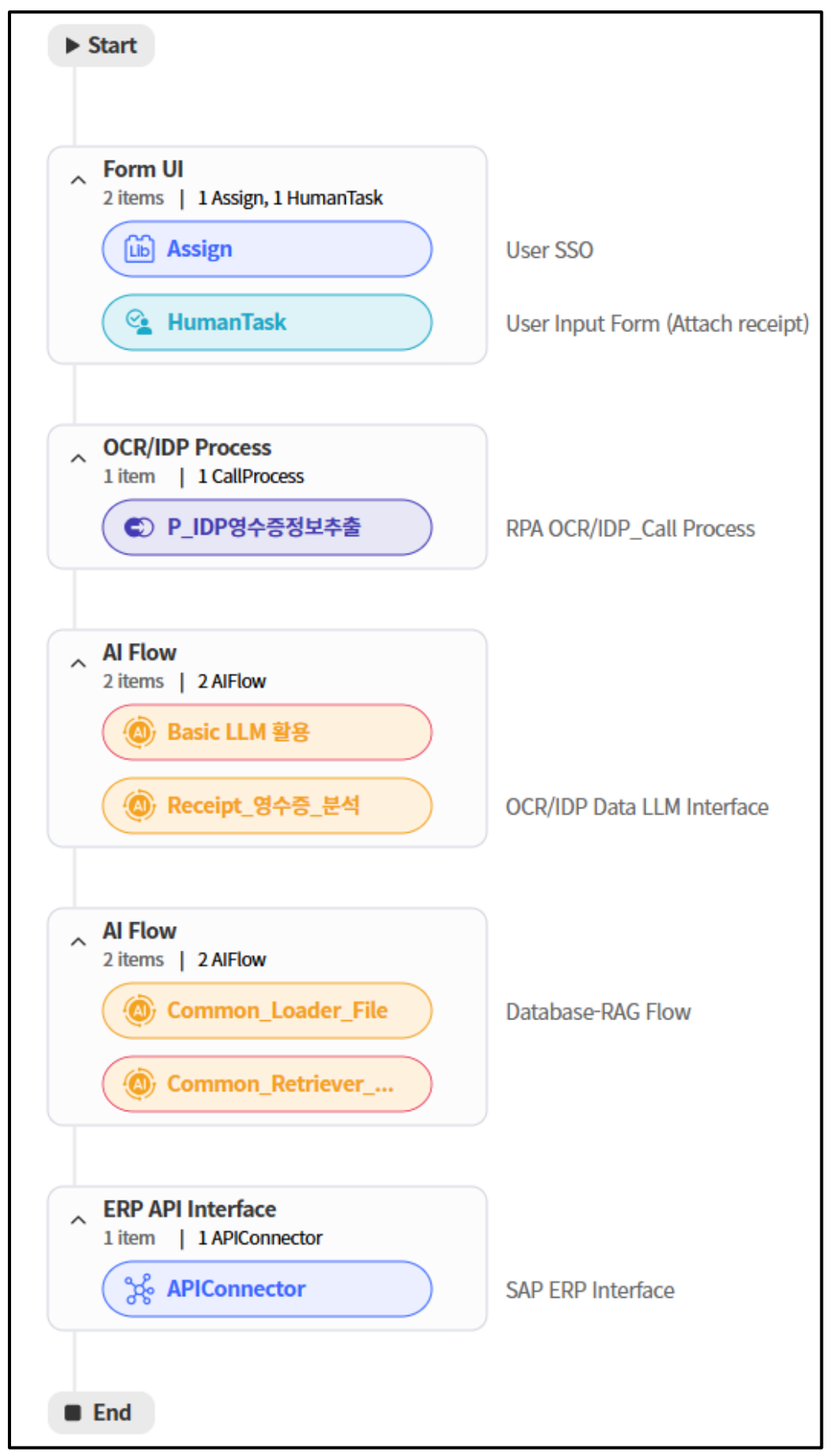

The moment of human intervention occurs only once, but that decision enables hundreds of automated processes, suggesting that human intellectual judgment can be effectively integrated into the system, achieving sustained automation efficiency without repeated interventions. This can be considered a significant advancement overcoming the limitations of existing RPA, as it goes beyond the automation of simple repetitive tasks and extends to the automation of even “complex processes requiring human intervention”.

This case suggests the possibility of extending a similar E2E automation model to other complex business areas that require human judgment and exception handling, such as accounting, human resources, and procurement, beyond the specific domain of financial expense processing. It is expected to contribute to creating an environment where employees can focus on more creative and strategic tasks by maximizing operational efficiency through hyper-automation.

# 5. Evaluation Results

In this chapter, we analyze the effects of introducing end-to-end automation for financial expense processing in a generative AI and OCR/IDP-based Automation Agent system utilizing Brity Automation. The analysis is divided into quantitative and qualitative aspects. Additionally, we present potential limitations that may arise during the introduction process and propose future improvement plans to overcome these challenges and further enhance the system.

## 5.1. Quantitative Effects

The most direct effect of introducing an automation system is the improvement in work efficiency and cost reduction. As a result of the pilot implementation, the evaluation of 1,448 paper receipts (png, jpg, etc., image files) out of the total card expenses incurred over a month showed the results as shown in Tables 4 and 5.

**Table 4**
**Confusion Matrix Validation Target**

| Classification | Normal | | Error | |
|---|---|---|---|---|
| Verification Target | TP (Agreement/Agreement) | TN (Rejected/Rejected) | FP (Rejected /Approval) | FN (Agreement/Rejection) |
| 1,448 items | 1,073 | 133 | 0 | 242 |

**Table 5**
**Quantitative Evaluation Metrics**

| Performance Metrics | Calculation Formula | Result |
|---|---|---|
| Accuracy | *(TN+TP)/(TN+FN+FP+TP)* = (133+1,073)/(133+242+0+1,073) | 0.83 |
| Precision | *TP/(TP + FP)* = 1,073/(1,073 + 0) | 1 |
| Recall | *TP/(TP + FN)* = 1,073/(1,073 + 242) | 0.82 |
| F1 Score | *2 / ((1/ Precision )+(1/ Recall ))* = 2 / (1/1 + 1/0.82) | 0.90 |

An F1 score of 0.90 indicates that the model is performing well in both precision and recall, demonstrating a well-balanced trade-off between the two. This balance reflects a model that prioritizes both the accuracy of positive predictions and minimizing false negatives.

In addition, the expected quantitative effects based on general intelligent automation adoption cases are as follows:

1. **Significant Reduction in Task Processing Time:** In the case of Company S, the traditional manual process for handling financial and expense management tasks involved an average of approximately 1,450 cases per month, each requiring more than 24 hours to complete. The implementation of Intelligent IDP, which enables automatic information extraction, policy-based categorization, exception handling support through AI Flow, and an increasing automation rate via a learning-based Automation Agent, led to a remarkable reduction in overall processing time. The time required per case decreased from several minutes to just a few tens of seconds, resulting in a monthly time savings of over 80% to 90%. This substantial efficiency gain allows financial personnel to shift their focus from repetitive administrative tasks to more analytical and strategic functions.
2. **Cost Reduction in Operations:** Reducing the time required to complete tasks directly leads to a decrease in labor costs. By minimizing the manpower allocated to specific tasks or enabling the same workforce to handle more tasks, it results in tangible cost savings. Additionally, it can also reduce indirect costs such as those incurred from correcting errors caused by manual work, managing paper documents, and other related expenses.
3. **Increased processing capacity and scalability:** Automated systems can operate 24/7, allowing for flexible handling of increased workloads. They effectively distribute the concentrated expense processing tasks that occur at the end of the month or year and provide scalability to respond to increased workloads due to business growth without the need for additional personnel.
4. **Error Rate Reduction:** Accurate data extraction through IDP and system-based verification processes can significantly reduce human errors (such as typos, omissions, etc.) that may occur during manual input. This enhances the accuracy of data, improving the reliability of financial reporting, and reduces the time and cost spent on error correction.
5. **Improving compliance rates:** The built-in policy database and AI-based validation logic ensure that all expense processing is consistently carried out in accordance with predefined company regulations and policies. This leads to enhanced internal controls and increased ease of responding to audits.

### 5.2. Qualitative Effects

The introduction of E2E automation brings about a variety of qualitative benefits that positively impact organizational culture and work methods.

1. **Improving Work Accuracy and Consistency:** Automated processes handle tasks based on consistent criteria without deviations caused by subjective human judgment or individual conditions, thereby enhancing the overall quality and accuracy of work. In particular, as AI continuously improves its decision-making standards through learning, the accuracy of handling complex exception situations can gradually increase.
2. **Increased Employee Satisfaction and Work Engagement:** Financial professionals move away from simple, repetitive receipt processing tasks and take on more advanced roles, such as reviewing AI-generated analysis results and making final decisions. This enhances job satisfaction and creates an environment where employees can focus on more creative and valuable work. While human intervention occurs only once, the judgment made enables hundreds of automations, highlighting the value of employee involvement in elevating the importance of their input.
3. **Enhancing Data-Driven Decision Support:** The vast amount of structured expense data accumulated throughout the E2E automation process can provide valuable insights for various decision-making processes, such as cost pattern analysis, budget management, and strengthening internal controls. For instance, it can be utilized to identify trends of excessive spending in specific departments or categories or to detect regulatory violations at an early stage.
4. **Rapid Response to Change and Enhanced Flexibility:** Through AI Flow and Automation Agents, the system can quickly learn and adapt to new types of items or changes in policies. This provides the flexibility to respond swiftly to changes in the business environment or new regulatory requirements.

### 5.3. Limitations in the Introduction Process and Future Improvement Plans

Intelligent process automation using generative AI and OCR/IDP-based automation agents can contribute to improving business efficiency for companies. However, there are various limitations in the actual implementation process. First, high costs and prolonged time may be incurred during the initial setup, which can be mitigated through the use of cloud-based SaaS, phased implementation strategies, and the utilization of pre-trained models. Second, the performance of IDP and AI models heavily depends on the availability of high-quality training data, necessitating continuous accumulation of learning data through data augmentation techniques and the Human-in-the-loop approach. Third, since the decision-making process of generative AI operates as a black box, ensuring trustworthiness and explainability in judgments is required. This can be achieved by presenting decision-making grounds, linking relevant policies, and introducing XAI technologies to enhance user acceptance. Fourth, the lack of technical acceptance among organizational members is a critical task in change management, requiring user education, interface improvements, and system design based on active participation from business units. Fifth, since automated systems handle expense data containing sensitive information, technical and managerial security measures such as data encryption, access control, and regular security audits are essential. Lastly, as policies and environments for automated tasks continuously evolve, flexible system architecture design, establishment of version management systems, and self-learning-based maintenance automation are necessary in the long term.

By being aware of these limitations in advance and preparing appropriate improvement measures, companies can increase the likelihood of successfully introducing intelligent automation systems and maximize their effectiveness.

## 6. Discussion and Conclusion

***Summary and Conclusion***

This study deeply analyzed and presented a method for implementing end-to-end automation of financial expense processing tasks by combining generative AI and OCR/IDP technologies with Automation Agents, focusing on cases utilizing Brity Automation. The conclusions drawn from the study are as follows:

Firstly, the existing financial expense management process had various issues for automation implementation, including high reliance on manual work, frequent exceptions, and difficulties in ensuring data accuracy. Simply using RPA was insufficient to address this complexity. To overcome these challenges, it was confirmed that accurate document recognition through IDP, intelligent exception handling and decision support using generative AI, and the introduction of an Automation Agent that collaborates between humans and AI are essential.

Secondly, through the case analysis of Brity Automation, it was demonstrated that a four-step approach—'document recognition' utilizing IDP, 'policy-based classification' based on a database, 'exceptional AI inquiry' through AI Flow (LLM integration), and 'systematic automation of user judgment' via Automation Agent—can effectively implement E2E automation in financial expense processing. Notably, the Human-in-the-loop mechanism, where the system learns from human final judgments to continuously improve automation performance, served as a key element in intelligent automation.

Third, the proposed automation system was found to bring not only quantitative benefits such as reduced processing time, cost savings, and decreased error rates, but also various qualitative benefits such as improved work accuracy, increased employee satisfaction, and enhanced support for data-driven decision-making. This can contribute to enhancing the productivity and competitiveness of the company.

Fourth, it emphasized the importance of recognizing the limitations and challenges in the process of introducing intelligent automation systems, such as initial construction costs, data quality assurance, explainability of AI models, change management, and security, and preparing appropriate improvement measures accordingly.

In conclusion, the organic integration of generative AI, OCR/IDP, databases, and automation agents enables successful E2E automation even in complex and exception-prone areas of work, such as financial expense processing. This goes beyond mere task replacement and holds the potential to fundamentally transform corporate operational methods. The Brity Automation case serves as a significant empirical evidence demonstrating how such technological integration can lead to tangible business value.

### *Implications and Contributions of the Study*

This study has the following implications and contributions.

From an academic perspective, first, this study presents a concrete framework and practical case studies on how cutting-edge AI technologies, such as generative AI, can be integrated with traditional RPA and IDP to automate complex enterprise business processes. Second, by emphasizing the importance and implementation strategies of effective human-AI collaboration models (Human-in-the-loop) in the end-to-end automation process, it contributes to the field of intelligent automation research.

Practical implications include: First, providing a practical introduction strategy, technical configuration plan, expected effects, and potential challenges for companies considering financial expense processing automation. Second, through case analysis based on actual solutions such as Brity Automation, it offers specific information that companies reviewing the construction or introduction of similar systems can use for benchmarking. Third, it demonstrates the possibility of expanding the scope of automation beyond simple repetitive tasks to areas requiring human judgment, thereby presenting the potential for the spread of intelligent automation to other business domains.

### *Future Research Directions and Suggestions for Expansion*

This study has demonstrated the feasibility of end-to-end automation for financial expense processing tasks using generative AI and IDP-based automation agents.

This study demonstrated the possibility of E2E automation of financial expense processing tasks by combining generative AI and OCR/IDP technologies with Brity Automation. Future research can be expanded in the following directions:

First, the automation framework of this study has high applicability to various industries such as finance, manufacturing, healthcare, and public sectors, as well as to non-routine document processing and exception-prone tasks such as contract management, customer service, and legal review. Therefore, there is a need for case studies and effectiveness analysis reflecting the characteristics of each field.

Second, advanced research is required to enhance the domain-specific learning and precision of generative AI models. Additionally, integrating XAI (Explainable AI) technology to explain AI decision-making can improve system transparency and user acceptance.

Third, research on an integrated approach is needed to discover automation target processes based on actual work log data through process mining technology and to quantitatively analyze and optimize the operational effectiveness of automation systems.

Fourth, as intelligent automation spreads, it is necessary to analyze ethical and social impacts such as job restructuring, data privacy, and algorithmic bias, and to conduct research on policy and institutional measures to mitigate these issues.

Lastly, research is needed to explore the potential of multimodal AI-based automation technologies that can simultaneously understand and process various types of data beyond text and images, such as voice and video.

Such follow-up studies are expected to enhance the practicality of intelligent automation technologies while making significant contributions to corporate digital transformation and sustainable work innovation.